\documentclass{article}

\PassOptionsToPackage{numbers, compress}{natbib}
\usepackage[main, final]{neurips_2026}

\usepackage[utf8]{inputenc} 
\usepackage[T1]{fontenc}    
\usepackage{hyperref}       
\usepackage{url}            
\usepackage{booktabs}       
\usepackage{amsfonts}       
\usepackage{nicefrac}       
\usepackage{microtype}      
\usepackage[table]{xcolor}        
\usepackage{graphicx}
\usepackage{amsmath}
\usepackage{subcaption}
\usepackage{multirow}
\usepackage{enumitem}
\usepackage{tikz}
\usepackage{amsmath,amssymb}
\usepackage{fontawesome5}
\usepackage{subcaption}
\usepackage{tcolorbox}
\usetikzlibrary{
  positioning,
  arrows.meta,
  calc,
  shapes.geometric
}
\definecolor{encblue}{HTML}{A8D4E6}
\definecolor{encborder}{HTML}{3A7FA8}
\definecolor{headpeach}{HTML}{F0C89A}
\definecolor{headborder}{HTML}{C47A20}
\definecolor{lossred}{HTML}{C0392B}
\definecolor{arrgray}{HTML}{555555}
\definecolor{titlegray}{HTML}{333333}
\definecolor{sublabelgray}{HTML}{888888}

\belowdisplayskip \abovedisplayskip

\newlength{\sectionReduceTop}
\newlength{\sectionReduceBot}
\newlength{\subsectionReduceTop}
\newlength{\subsectionReduceBot}
\newlength{\abstractReduceTop}
\newlength{\abstractReduceBot}
\newlength{\captionReduceTop}
\newlength{\captionReduceBot}
\newlength{\subsubsectionReduceTop}
\newlength{\subsubsectionReduceBot}

\newlength{\eqnReduceTop}
\newlength{\eqnReduceBot}

\newlength{\horSkip}
\newlength{\verSkip}

\newlength{\figureHeight}
\newcommand{\phyprobe}{\text{\textsc{PhyProbe}}}

\title{\phyprobe: Rethinking Physical Consistency Evaluation in Generated Videos}

\author{
\begin{tabular}{c}
Max Ku$^{123}$\thanks{Part of the work done during internship at NVIDIA Cosmos Lab}  \quad Jiaojiao Fan$^{1}$ \quad Zekun Hao$^{1}$ \quad Francesco Ferroni$^{1}$ \\
Heng Wang$^{1}$ \quad Wenhu Chen$^{23}$ \quad Ming-Yu Liu$^{1}$ \quad Prithvijit Chattopadhyay$^{1}$ \\
\\
$^{1}$NVIDIA \quad $^{2}$University of Waterloo \quad $^{3}$Vector Institute
\end{tabular}
}

\begin{document}

\maketitle

\begin{abstract}
Evaluating the physical consistency of generated videos remains a fundamental challenge. Existing approaches rely on off-the-shelf vision-language models, which can often be myopic to physical dynamics, or fine-tuned evaluators trained on human annotations, which overfit to dataset-specific cues and fail to generalize. A key challenge is that existing supervision sources provide either relative ordering or absolute scores, but not both reliably and consistently across varied settings.
To this end, we introduce \phyprobe{}, an evaluator that extracts features from a frozen pretrained spatio-temporal encoder and maps them to a scalar physical consistency violation score via a lightweight scoring head. \phyprobe{} is trained through a unified objective combining pairwise ranking, regression on noisy scalar annotations, and anchor-based calibration over a curated set of heterogeneous supervision sources. Experiments show that \phyprobe{} outperforms prior methods on most pairwise benchmarks spanning real–generated and generated–generated pairs under varying correspondence, with the largest gains in no-correspondence and generated–generated settings where existing fine-tuned evaluators degrade sharply. \phyprobe{} achieves strong correlation with human judgments, with close agreement between rank-based and linear metrics, indicating that scores are both well ordered and anchored to a stable [0, 1] scale. Further, despite being trained on supervision indicative of physical consistency, without explicit general-preference labels, \phyprobe{} also performs competitively on human preference benchmarks: consistent with the observation that physics violations are entangled with broader quality degradations.
\end{abstract}
\section{Introduction}
\label{sec:intro}

\begin{figure}[t]
    \centering
    \includegraphics[width=0.95\linewidth]{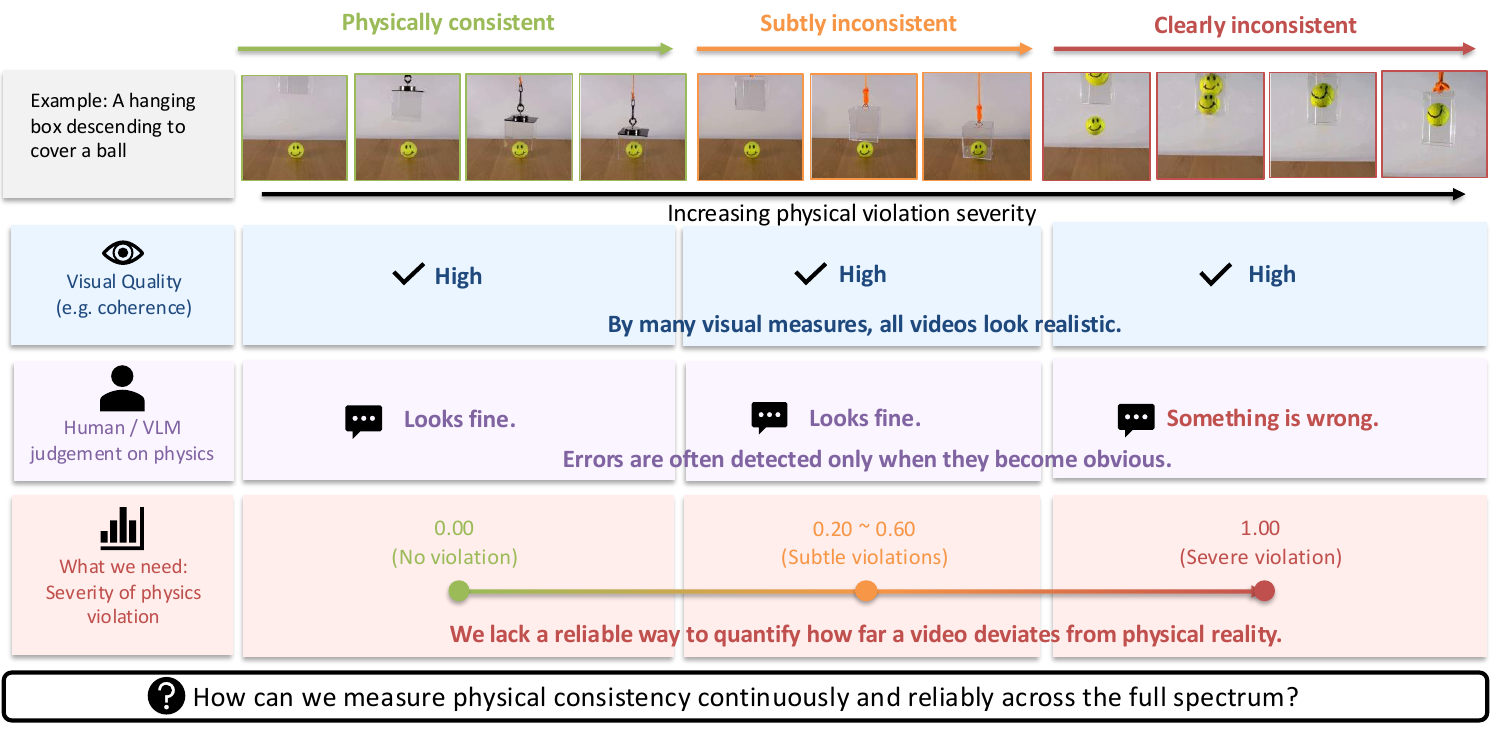}
    \caption{
    \textbf{Challenges in evaluating physical consistency.} Generated videos can be visually realistic across the board, while their physical consistency varies from fully consistent to severely violated. Human perception and current VLM-based evaluators can be thresholded and inconsistent, making it difficult to capture the continuous spectrum of violation severity. This motivates our goal: learning a bounded, continuous measure of physical consistency violation.
    }
    \label{fig:teaser}
\end{figure}

Recent advances in video generation~\citep{yang2025cogvideoxtexttovideodiffusionmodels, hacohen2024ltxvideorealtimevideolatent, wan2025wanopenadvancedlargescale, nvidia2025cosmosworldfoundationmodel} have made considerable progress across several dimensions: from producing videos of high visual quality~\citep{openai2024sora, hailuoai}, temporally coherent motion~\citep{klingteam2025klingomnitechnicalreport, LumaRay3}, to native multimodal support~\citep{seedance2026seedance20advancingvideo, googleveo3report2025, hacohen2026ltx2efficientjointaudiovisual}. Yet one critical dimension remains stubbornly difficult to evaluate: \textbf{physical consistency}. A generated video can exhibit sharp textures and smooth motion while depicting objects that defy gravity, pass through solid surfaces, or spontaneously change form~\citep{bai2025impossiblevideos, lin2025brokenvideosbenchmarkdatasetfinegrained}. But improving physical consistency requires first being able to measure it, and reliable measurement remains an open problem~\citep{riochet2020intphysframeworkbenchmarkvisual,bansal2024videophyevaluatingphysicalcommonsense}. The difficulty here is not in detecting visual quality degradations -- often, there may not be any —- but in measuring how far the underlying dynamics in the generated video depart from physical reality. 
Crucially, simply identifying whether a video is physically plausible or not is insufficient. Not all physical violations are equal: an object drifting slightly off a surface is qualitatively different from an object exploding into fragments mid-air. A useful evaluator for \textit{physical consistency} must quantify the ``egregiousness'' of a violation: producing a continuous, consistently scaled score that reflects severity, not just presence. This frames physical consistency evaluation as a measurement problem, and measurement demands precision, stability, and scale, as illustrated in Figure~\ref{fig:teaser}. Measurement must operate over a suitable representation of video dynamics; we argue that this representation should capture spatiotemporal structure directly rather than relying on semantic proxies such as language descriptions.

Existing approaches have made meaningful progress, but suffer from limitations in both evaluator design and supervision. Off-the-shelf VLMs~\citep{jiang2024genaiarenaopenevaluation, li2024genaibenchevaluatingimprovingcompositional, google_gemini3} increasingly serve as evaluators through semantic reasoning, yet remain myopic to physical dynamics and require carefully crafted, context-dependent prompts to expose violations. Fine-tuned VLM evaluators~\citep{bansal2024videophyevaluatingphysicalcommonsense, bansal2025videophy2challengingactioncentricphysical, he2024videoscorebuildingautomaticmetrics, he2025videoscore2thinkscoregenerative,motamed2025travlrecipemakingvideolanguage} and detection-based approaches~\citep{wang2025phydetexdetectingexplainingphysical, gao2025davidxr1detectingaigeneratedvideos, physionlabs2026galileo0} improve performance through supervision on human annotations, but still operate largely on semantic proxy signals, relying on dataset-specific shortcuts rather than modeling underlying temporal dynamics. As a result, their performance often degrades under distribution shift. 
Even though human annotations remain the most reliable ground truth for  physical consistency, they can be noisy at times: observers tend to detect only salient violations while absolute ratings drift as generative quality improves~\citep{nightingale2017identify,nightingale2019detect,nightingale2022ai}. This motivates an evaluator that learns from such annotations without being limited by their precision. Moreover, physical inconsistencies often co-occur with other visual artifacts, causing annotators to respond to overall perceptual quality rather than physical dynamics specifically. Consequently, absolute ratings often reflect the prevailing data distribution at annotation time rather than a stable notion of physical consistency, and evaluators trained on such annotations inherit the resulting noise and drift (Details in Section~\ref{sec:results}).

\par \noindent
\textbf{Desiderata.} These limitations motivate us to establish a clear desiderata of what a reliable physical consistency evaluator should achieve. (1) Reliable relative orderings: even if absolute scoring is imperfect, consistent "video A has more physical consistency violations than video B" judgments are sufficient for measurement. (2) Anchored scores on a fixed reference scale, so that absolute values remain comparable across datasets and time periods rather than drifting with the data distribution. (3) Operate on a single video at inference time, without requiring a paired reference or prompt alignment, since many physical violations are identifiable from temporal dynamics alone~\citep{yasuda2021physical, margoni2024violation}. 

\par \noindent
\textbf{Contributions.} To this end, we introduce \phyprobe{}, a video evaluator for physical consistency. Given a generated video, \phyprobe{} extracts features from a frozen pretrained spatio-temporal encoder and maps them through a lightweight scoring head to a ``violation score'', $s(v)$, indicative of the degree of physical consistency violation. We find that the key to learning a meaningful scoring function like \phyprobe{} lies in the following:

\begin{enumerate}[leftmargin=*, align=left]
    \item \textbf{Strong spatio-temporal representations.} We show that frozen pretrained spatio-temporal encoders~\citep{assran2025vjepa2selfsupervisedvideo, bolya2025perceptionencoderbestvisual}, trained through self-supervised or weakly-supervised objectives on large-scale video data, provide a foundation on which a lightweight scoring head can map video representations to a scalar violation score, without requiring VLMs or language-mediated reasoning.
    \item \textbf{Curating rich supervision from heterogenous sources.} We reorganize diverse supervision sources into a unified training framework: real vs. generated comparisons (with and without semantic correspondence)~\citep{zhou2025paibenchcomprehensivebenchmarkphysical}, real vs. real pairs for consistency and calibration~\citep{kay2017kineticshumanactionvideo,carreira2022shortnotekinetics700human}, synthetic failure cases~\citep{bai2025impossiblevideos,lin2025brokenvideosbenchmarkdatasetfinegrained}, and human-annotated plausibility judgments~\citep{bansal2025videophy2challengingactioncentricphysical,he2025videoscore2thinkscoregenerative}. We organize these into three forms of supervision:
    \begin{enumerate}
        \item Pairwise Ranking Supervision: Stating "video A is more physically violated than video B" is more noise-tolerant than assigning absolute scores, providing stable relative ordering.
        \item Regress to Noisy Scalar Annotations: Where available, human-annotated severity scores provide approximate magnitude information that grounds the ordering in absolute values.
        \item Anchoring Violation Scores: Real videos map to 0, clearly violated videos map to 1, stabilizing the absolute scale across datasets.
    \end{enumerate}
    \item \textbf{Unified training objective for scoring.} The three supervision signals are combined into a single objective that decomposes the measurement problem into ranking (Bradley-Terry), magnitude (regression), and scale calibration (anchoring) losses, each addressed by the data-source best suited for it. \phyprobe{} trains only 1M parameters on top of a frozen 2B-parameter encoder, and is an effective evaluator for physical consistency.

\end{enumerate}

We extensively evaluate \phyprobe{} across diverse settings to assess preference accuracy, human alignment, and generalization. On pairwise benchmarks spanning real versus generated and generated versus generated comparisons under varying correspondence levels~\citep{wang2025phydetexdetectingexplainingphysical,bansal2025videophy2challengingactioncentricphysical, he2025videoscore2thinkscoregenerative,motamed2025travlrecipemakingvideolanguage}, \phyprobe{} maintains strong performance even in no correspondence and generated versus generated settings where prior evaluators are inferior. It also shows moderate correlation on VideoPhy2~\citep{bansal2025videophy2challengingactioncentricphysical} and strong correlation on VideoFeedback2~\citep{he2025videoscore2thinkscoregenerative} human judgment test sets, indicating well ordered and stably anchored scores. Despite being trained on supervision indicative of physical consistency, without explicit general-preference labels, \phyprobe{} further generalizes to broader human preference benchmarks~\citep{xu2026visionrewardfinegrainedmultidimensionalhuman, rapidata_hailuo02_marey_i2v_2025, rapidata_seedance1pro_i2v_2025, liu2025improvingvideogenerationhuman}, suggesting that representations for detecting physical violations capture broader quality signals.
\section{Related Work}
\label{sec:related_works}

\begin{figure}[t]
    \centering \includegraphics[width=0.85\linewidth]{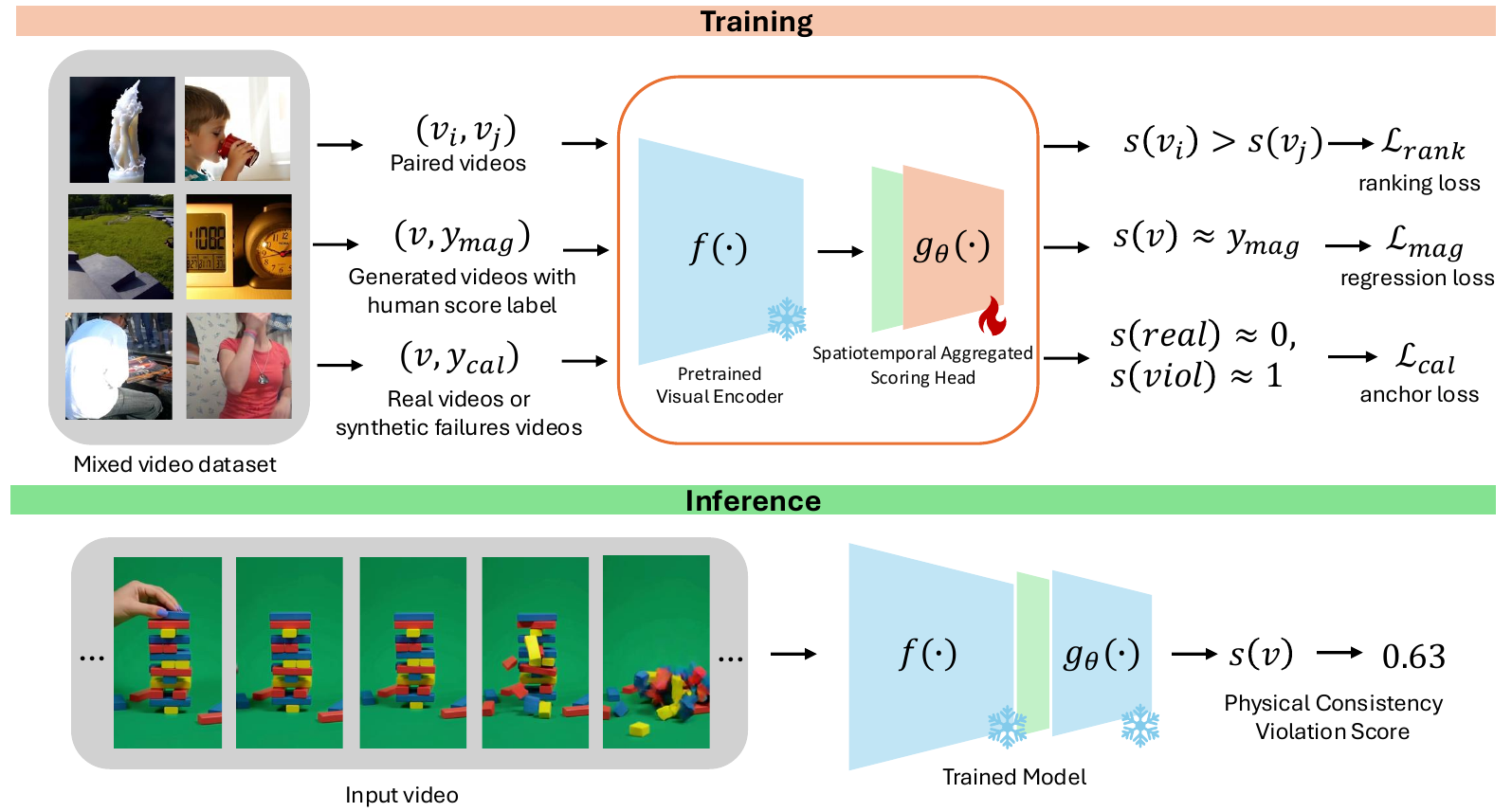}
    \caption{\textbf{Overview of \phyprobe{}.} Given a video, we extract a learned physical representation using a pretrained spatiotemporal visual encoder, followed by a lightweight scoring head to predict a continuous violation score $s(v)$. Training combines three complementary supervision signals: (i) pairwise ranking on video pairs $(v_i, v_j)$ to model relative violation, (ii) regression on human-annotated scores $y_{\text{mag}}$, and (iii) anchoring with calibration targets to enforce absolute scale (e.g., $s(\text{real})=0$, $s(\text{viol})=1$). At inference time, the model operates on a single video without requiring pairwise comparison, producing a continuous estimate of physical violation severity.}
    \label{fig:ours_method}
\end{figure}

\textbf{Intuitive Physics and Human Perception.}
Understanding physical plausibility has long been studied in cognitive science under the framework of intuitive physics, which investigates how humans perceive and reason about physical interactions~\citep{WeihsEtAl2022InfLevel, BAILLARGEON1985191, voetrigger2022, elecphy2024, joseph2026interpretingphysicsvideoworld}. Prior work suggests that human judgments on visual abnormalities are often coarse and thresholded, with observers reliably detecting only salient violations while overlooking subtle inconsistencies~\citep{nightingale2017identify,nightingale2019detect,nightingale2022ai}. These findings indicate that human perception does not provide a precise or stable measure of physical correctness. Rather than modeling human reasoning itself, our work focuses on how such judgments are operationalized in evaluation, motivating the need for measurement frameworks that do not directly depend on unstable human perception. This further motivates evaluation approaches that avoid directly modeling human perception, and instead rely on more stable representation-level signals of physical dynamics.

\textbf{Video Physical Benchmarks and Datasets.}
A wide range of benchmarks have been proposed to evaluate physical plausibility in video generation, including Physics-IQ, PhyGenBench, WorldModelBench, Physion, and IntPhys2 etc.~\citep{motamed2025generativevideomodelsunderstand_physicsiq, meng2024worldsimulatorcraftingphysical, li2025worldmodelbenchjudgingvideogeneration, bear2022physionevaluatingphysicalprediction, bordes2025intphys2benchmarkingintuitive, chen2025physicalcoherencebenchmarkevaluating, tragoudaras2026evaluatingnewtonianmechanicsvideo}. More recent datasets such as ImpossibleVideos~\citep{bai2025impossiblevideos} and BrokenVideos~\citep{lin2025brokenvideosbenchmarkdatasetfinegrained} focus specifically on failure cases in generated videos. Many of these benchmarks rely on human annotations, either through pairwise preferences or scalar ratings, which are increasingly used to train automated evaluators~\citep{bansal2024videophyevaluatingphysicalcommonsense, bansal2025videophy2challengingactioncentricphysical, he2024videoscorebuildingautomaticmetrics, he2025videoscore2thinkscoregenerative, motamed2025travlrecipemakingvideolanguage, wang2025physcorrdualrewarddpophysicsconstrained}. However, such annotations are inherently unstable: physical inconsistencies often co-occur with rendering artifacts and compositional errors, causing annotators to respond to overall perceptual quality rather than physical dynamics specifically. Moreover, as generative models improve, the annotation scales may drift over time and become implicitly calibrated to the prevailing data distribution rather than a fixed notion of physical consistency. These limitations highlight the difficulty of isolating physical consistency as a distinct evaluation signal.

\textbf{Video Evaluators for Physical Consistency.} Recent approaches increasingly employ VLM-based evaluators trained on human annotations to assess physical plausibility in generated videos~\citep{he2024videoscorebuildingautomaticmetrics, he2025videoscore2thinkscoregenerative, motamed2025travlrecipemakingvideolanguage}, such as VideoPhy2~\citep{bansal2024videophyevaluatingphysicalcommonsense, bansal2025videophy2challengingactioncentricphysical}. Other approaches fine-tune VLMs to detect or localize specific physical violations~\citep{wang2025phydetexdetectingexplainingphysical, gao2025davidxr1detectingaigeneratedvideos, physionlabs2026galileo0}. While effective on benchmark tasks, these methods fundamentally operate on semantic proxy signals and annotation patterns rather than directly modeling physical dynamics, making them sensitive to training data coverage and prone to degradation under distribution shift or subtle out-of-distribution violations. In contrast, our work formulates physical consistency evaluation as a representation-level measurement problem over temporal dynamics.

\section{\phyprobe{}}
\label{sec:method}
\phyprobe{} is designed to measure physical consistency in generated videos, mapping a video $v$ to a scalar violation score $s(v)$ that indicates the extent of physical consistency violation. To be effective, such an evaluator must be able to: (1) quantify violation severity through reliable relative ordering across videos and, (2) provide absolute scores on a stable, anchored scale to enable comparison across settings. We describe how \phyprobe{} achieves this through its architecture, data curation pipeline, and training objective.

\begin{table*}[ht!]
\small
\centering
\caption{
\textbf{Training datasets used for \phyprobe{}.} Each dataset contributes to different supervision signals: pairwise preference learning (rank), regression (mag), and anchor-based calibration (cal). During training, we sample data from different sources using dataset-level weights to balance their contributions, preventing large datasets from dominating the learning signal.
}
\label{tab:training_data}
\begin{tabular}{l r l l c c c}
\toprule
\textbf{Training Dataset} & \textbf{Pairs} & \textbf{Type} & \textbf{Correspondence} 
& \textbf{Rank} & \textbf{Mag} & \textbf{Cal} \\
\midrule
PAI-Bench~\citep{zhou2025paibenchcomprehensivebenchmarkphysical} & 42K & R-G & Image + Text & \checkmark & & \checkmark \\
VideoPhy2~\citep{bansal2025videophy2challengingactioncentricphysical} & 19K & G-G & Action / Text & \checkmark & \checkmark & \\
VideoFeedback2~\citep{he2025videoscore2thinkscoregenerative} & 31K & G-G & None & \checkmark & \checkmark & \\
BrokenVideos~\citep{lin2025brokenvideosbenchmarkdatasetfinegrained} & 14K & G-G & None & \checkmark & \checkmark & \\
ImpossibleVideos~\citep{bai2025impossiblevideos} & 27K & R-G & None & \checkmark & & \checkmark \\
TRAVL~\citep{motamed2025travlrecipemakingvideolanguage} & 98K & R-G & None & \checkmark & & \checkmark \\
Kinetics-700~\citep{carreira2022shortnotekinetics700human} & 140K & R-R & Action & \checkmark & & \checkmark \\
\midrule
Total & 371K \\
\bottomrule
\end{tabular}
\end{table*}

\par \noindent
\textbf{Architecture.} Given a video $v$ (generated or real), we extract video representations using a frozen pretrained spatio-temporal visual encoder (we primarily use Perception 
Encoder~\cite{bolya2025perceptionencoderbestvisual}) to obtain a feature representation $\mathbf{z} = f(v)$, 
where $f(\cdot)$ denotes the frozen backbone. The input videos 
are preprocessed as fixed-length clips with frames $T$, 
resized to a fixed spatial resolution with the aspect ratio 
preserved via padding, followed by backbone-specific normalization. The resulting representation $\mathbf{z}$ is mapped to a scalar violation score via a lightweight learned scoring head $g_\theta(\cdot)$, i.e., $s(v) = g_\theta(\mathbf{z})$. In principle, \phyprobe{} is agnostic to the choice of pretrained backbone, and we evaluate multiple architectures in Section~\ref{sec:results}. To obtain a video-level descriptor, we aggregate per-frame backbone features using a statistics pooling operator that concatenates the temporal mean, standard deviation, and max across timesteps. This captures overall content, temporal variability, and salient events, providing a simple yet effective summary of the feature distribution while remaining lightweight and invariant to video length. The aggregated representation is then passed to a lightweight three-layer Multi-Layer Perceptron (MLP) with GELU activations that maps it to a scalar violation score. The design is intentionally simple and parameter-efficient, allowing the pretrained backbone to handle spatiotemporal modeling while the head focuses on learning a smooth, consistently scaled scoring function. We ablate alternative aggregation strategies in Section~\ref{sec:results_ablations}.

\par \noindent
\textbf{Data Curation.} We train \phyprobe{} on a curated dataset of approximately 
371K video pairs drawn from seven diverse sources. Each data source spans a range of pair types (real vs.\ generated comparisons~\citep{zhou2025paibenchcomprehensivebenchmarkphysical}, real vs.\ real pairs for consistency and calibration~\citep{kay2017kineticshumanactionvideo, carreira2022shortnotekinetics700human}, synthetic failure cases~\citep{lin2025brokenvideosbenchmarkdatasetfinegrained, bai2025impossiblevideos}, and human-annotated judgments~\citep{bansal2025videophy2challengingactioncentricphysical, he2025videoscore2thinkscoregenerative}) under varying degrees of semantic correspondence as shown in Table~\ref{tab:training_data}. This heterogeneity is deliberate: 
by exposing the model to comparisons with and without 
shared prompts or content, we discourage reliance on semantic 
shortcuts and encourage sensitivity to physical consistency 
itself. The videos cover a broad range of physical violation types: behavioral abnormalities such as object interpenetration, unrealistic fluid dynamics, implausible deformation, and erratic motion violating expected force and collision dynamics; time abnormalities including temporal jumps, motion reversal, and violations of causal ordering; objectness abnormalities such as objects suddenly appearing or disappearing, unexpected morphing, and inconsistent spatial attributes across frames; and perception abnormalities in the form of visual breakdown, local blurring, and style inconsistencies introduced by generative artifacts.

\par \noindent
\textbf{Learning Objectives.} We categorize the training supervision from these data sources into three complementary forms, each addressing a distinct aspect of the scoring problem.

\begin{enumerate}[leftmargin=*, align=left]
    \item \textbf{Ranking Supervision.} We construct video pairs from both 
    aligned sources (shared prompts or content) and unaligned 
    sources (no shared prompt), providing relative ordering 
    signals under varying semantic correspondence. Including 
    both prevents the model from being over-reliant on shared semantic correspondence. To recover a scalar score from pairwise preferences, we adopt the Bradley-Terry formulation~\cite{Bradley1952RankAO}, which treats $s(\cdot)$ as a latent severity score and models the probability that $v_i$ is more violated than $v_j$ as $\sigma(s(v_i) - s(v_j))$, where $\sigma$ is the logistic sigmoid. Given a pair $(v_i, v_j)$ from the ranking supervision dataset $\mathcal{D}_{\text{rank}}$, we define:
\begin{equation}
\mathcal{L}_{\text{rank}} = - \mathbb{E}_{(v_i, v_j, y_{ij})\sim\mathcal{D}_{\text{rank}}} 
\left[ y_{ij} \log \sigma(s(v_i) - s(v_j)) 
+ (1 - y_{ij}) \log \sigma(s(v_j) - s(v_i)) \right],
\label{eq:loss_rank}
\end{equation}
where $y_{ij} \in \{0, 0.5, 1\}$ indicates the relative preference between $v_i$ and $v_j$: 1 if $v_i$ is more violated, 0 if $v_j$ is, and 0.5 for real–real pairs that carry no preference, so the ranking term pushes $s(v_i)$ and $s(v_j)$ toward equality. 

    \item \textbf{Regression for Violation Magnitude.} Noisy scalar annotations provide 
    approximate violation severity. Different datasets use 
    varying scales and definitions, so for datasets with 
    discrete or ordinal ratings (e.g., VideoFeedback2~\cite{he2025videoscore2thinkscoregenerative}, 
    VideoPhy2~\cite{bansal2025videophy2challengingactioncentricphysical}), we apply monotonic transformations to a unified range while preserving relative ordering. When explicit scores are unavailable, we derive 
    approximate magnitudes from dataset-specific signals (e.g, spatial extent and duration of annotated errors in BrokenVideos~\cite{lin2025brokenvideosbenchmarkdatasetfinegrained}), treating them as noisy measurements. To model the severity of violations, given noisy scalar annotations from $\mathcal{D}_{\text{noisy}}$ we incorporate a regression objective using the Huber loss~\citep{Huber1964RobustEO}:
\begin{equation}
\text{Huber}(s, y) =
\begin{cases}
\frac{1}{2}(s - y)^2, & \text{if } |s - y| \leq \delta \\
\delta (|s - y| - \frac{1}{2}\delta), & \text{otherwise}
\end{cases}
\label{eq:loss_huber}
\end{equation}
\begin{equation}
\mathcal{L}_{\text{mag}} = 
\mathbb{E}_{v \sim \mathcal{D}_{\text{noisy}}} 
\left[ \text{Huber}(s(v), y_v) \right],
\label{eq:loss_mag}
\end{equation}

    \item \textbf{Anchor Loss for Scale Calibration.} Anchor samples represent 
    approximate endpoints of the violation spectrum: they are selected to represent approximate endpoints, including high-confidence real videos~\citep{kay2017kineticshumanactionvideo} (low violation; anchor toward $s \approx 0$) and synthetic failure cases~\citep{bai2025impossiblevideos} (high violation; anchor toward $s \approx 1$). Real-real pairs reinforce the near-zero baseline, and real-generated comparisons help establish relative scale 
    across domains. These provide weak but stable reference 
    points for resolving scale ambiguity across heterogeneous 
    sources (a curated $\mathcal{D}_{\text{anchor}}$ dataset with binary labels). To calibrate the score range, we introduce an anchor loss:
\begin{equation}
\mathcal{L}_{\text{cal}} = 
\mathbb{E}_{v \sim \mathcal{D}_{\text{anchor}}} \left[ 
\| s(v) - \mathbf{I(v)} \|_2^2 
\right],
\label{eq:loss_cal}
\end{equation}
where
$\mathbf{I}(v) = 
\begin{cases}
  0 & \text{if } v \in \mathcal{D}_{\text{anchor}}^{\text{plausible}} \\
  1 & \text{if } v \in \mathcal{D}_{\text{anchor}}^{\text{violated}}
\end{cases}$ based on the curated $\mathcal{D}_{\text{anchor}}$ samples.
\end{enumerate}



The overall objective is defined as:
\begin{equation}
\mathcal{L} =
\mathcal{L}_{\text{rank}} +
\lambda_{\text{mag}} \mathcal{L}_{\text{mag}} +
\lambda_{\text{cal}} \mathcal{L}_{\text{cal}},
\label{eq:loss_total}
\end{equation}

In practice, the $[0,1]$ violation interval should be viewed as a calibration target rather than a strict architectural constraint. While the regression and anchor losses encourage scores to concentrate near this range during training, the scoring head itself remains unconstrained and does not explicitly enforce bounded outputs. To obtain a stable and interpretable violation scale, we report $s(v) = \text{clip}(\tilde{s}(v), 0, 1)$ at inference time. Empirically, we observe that the learned score distributions already remain close to the target interval even without explicit constraints as shown in Table~\ref{tab:ablation_loss}.


\section{Experiment Results}
\label{sec:results}

\subsection{Experimental Setup}
\label{sec:results_setup}

\par \noindent
\textbf{Evaluation Benchmarks.} We evaluate \phyprobe{} predominantly on pairwise preference accuracy on multiple benchmarks spanning different levels of semantic correspondence and pair composition. The role and constitution of each evaluation dataset is summarized in Table~\ref{tab:eval_data_all} in Appendix~\ref{app:implementation_detail}. For pairwise physical consistency, we use four benchmarks:
ImplausibleBench~\cite{motamed2025travlrecipemakingvideolanguage} (image and text semantic correspondence with real-generated test
pairs), VideoPhy2~\cite{bansal2025videophy2challengingactioncentricphysical} (text correspondence with generated-generated pairs), PhyDetEx~\cite{wang2025phydetexdetectingexplainingphysical} (no semantic correspondence, with real-generated pairs), and VideoFeedback2~\cite{he2025videoscore2thinkscoregenerative} (no correspondence, with generated-generated pairs). Since absolute annotation scales vary across datasets, we convert scalar-annotated benchmarks into a pairwise format by ranking samples within each dataset, enabling consistent comparison alongside natively paired benchmarks. We report pairwise accuracy for ranking performance. To measure alignment with human-ratings (i.e. human correlation), we use VideoPhy2 and VideoFeedback2, both of which provide scalar physical plausibility annotations. We compute both Spearman's 
rank correlation~($\rho$) to assess ordering consistency and 
Pearson correlation~($r$) to evaluate linear alignment. We note that neither metric measures absolute calibration; we assess scale stability via the anchor ablation (Table~\ref{tab:ablation_loss}). To further measure \phyprobe{}'s \emph{alignment with human preference}, we evaluate on MonetBench~\cite{xu2026visionrewardfinegrainedmultidimensionalhuman}, RewardBench~\cite{liu2025improvingvideogenerationhuman}, and Rapidata~\cite{rapidata_hailuo02_marey_i2v_2025,rapidata_seedance1pro_i2v_2025}, and which assess broader video quality and preference alignment beyond physical consistency.

\par \noindent
\textbf{Baselines and Prior Methods.} We compare against five baselines spanning different evaluation paradigms: VideoPhy-2-AutoEval~\cite{bansal2025videophy2challengingactioncentricphysical}, a fine-tuned VLM evaluator for physical plausibility; VideoScore2~\cite{he2025videoscore2thinkscoregenerative}, a reward model trained on human preferences but also support physical and commonsense evaluation; V-JEPA Surprisal Score~\cite{assran2025vjepa2selfsupervisedvideo}, a self-supervised prediction-error metric based on V-JEPA2; finally Gemini-3.1-Flash-Lite~\cite{google_gemini3} and Gemini-3.1-Pro, off-the-shelf VLMs prompted zero-shot.

\subsection{Core Results}
\label{sec:results_core}

\begin{figure}[ht!]
    \centering
    \includegraphics[width=0.95\linewidth]{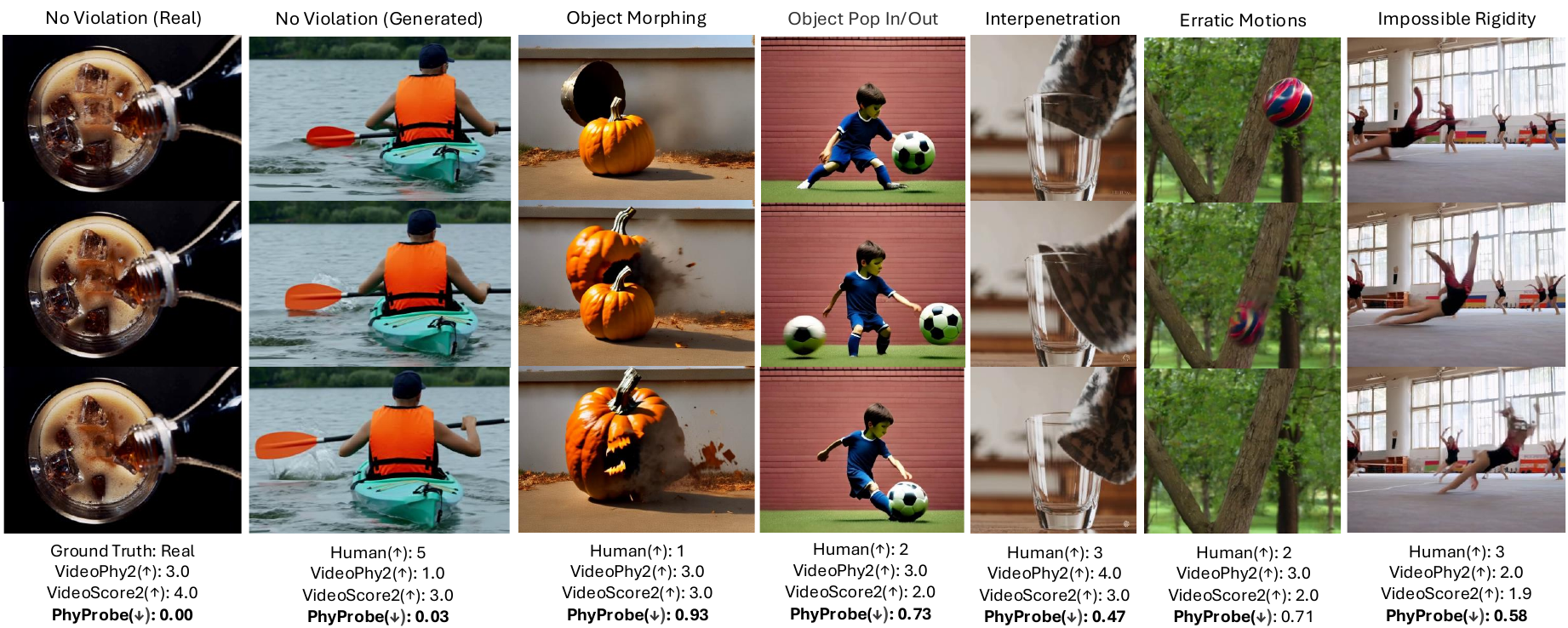}
    \caption{\textbf{Qualitative comparison of physical plausibility assessment across videos.} Each column shows representative frames, along with human annotations and scores from \phyprobe{} and prior physical evaluators~\citep{bansal2025videophy2challengingactioncentricphysical, he2025videoscore2thinkscoregenerative}. Arrows indicate whether the score is higher the better or vice versa.}
    \label{fig:ours_showcase}
\end{figure}
\begin{figure}[t!]
    \centering
    \includegraphics[width=0.95\linewidth]{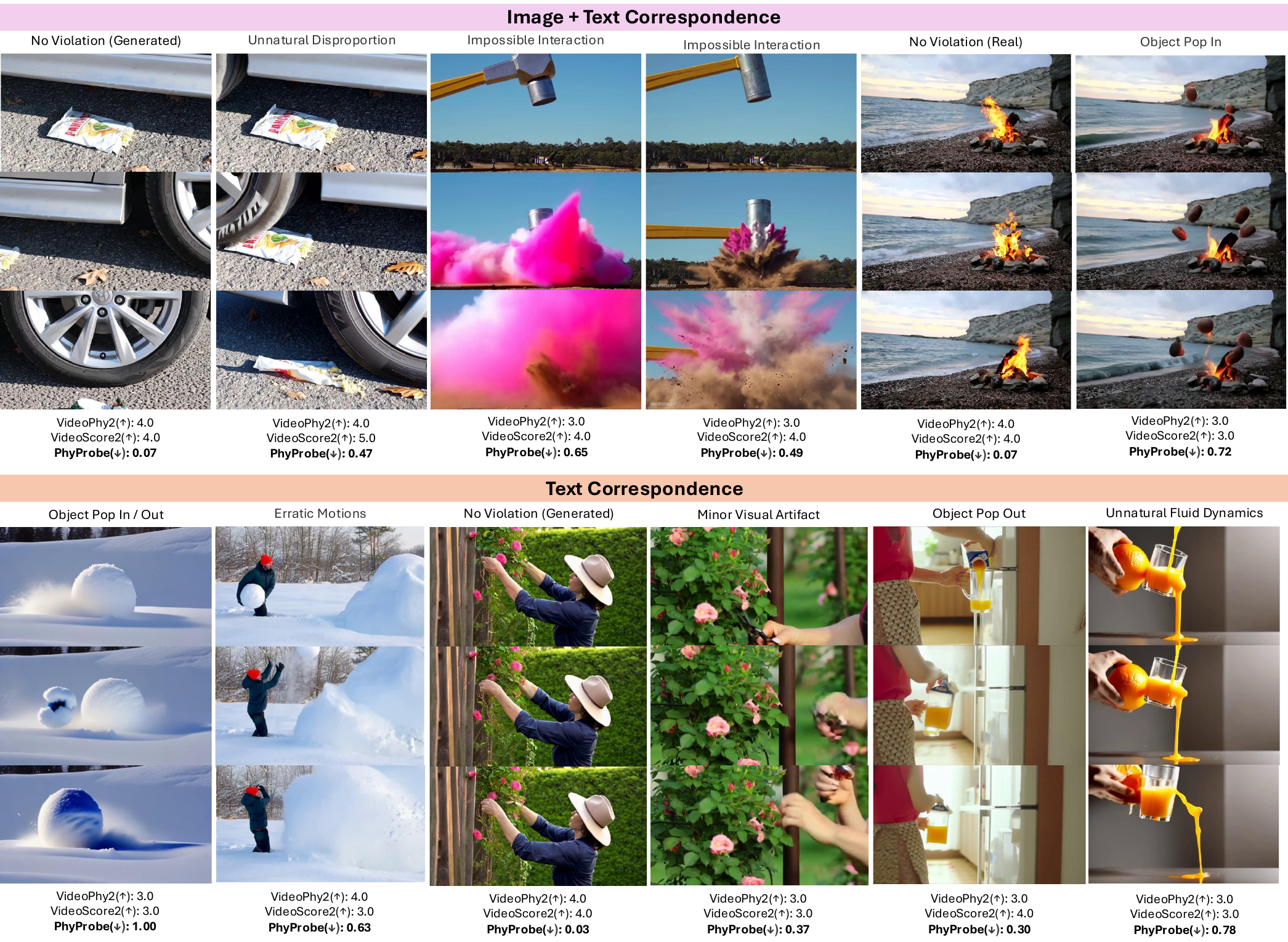}
    \caption{\textbf{Qualitative results across correspondence levels.}
    \phyprobe{} assigns low violation scores to videos with coherent object motion and interaction dynamics, while assigning higher scores to videos exhibiting implausible behaviors, effective across various correspondence level.
    }
    \label{fig:ours_qual_extra}
\end{figure}

\begin{table*}[t!]
\centering
\caption{\textbf{Core results on physical video benchmarks.} 
\textbf{Left Columns:} pairwise accuracy (\%) disentangled by pair type 
(real (R) vs generated (G)) and correspondence type (image+text, 
text, none). Benchmark$^p$ indicates the benchmark is reordered 
in pairwise manner without ties. We additionally report a weighted 
average across all pairwise benchmarks based on the number of pairs 
in each benchmark. 
\textbf{Right Columns:} correlation with human judgments on VideoPhy2 (VP2) 
and VideoFeedback2 (VF2). We report Spearman's rank correlation 
($\rho$) and Pearson correlation ($r$) to assess both ranking consistency and linear alignment. We additionally report Fisher-z averaged correlations across datasets, weighting each benchmark equally. We further studied the prior method behaviors in Appendix~\ref{app:d_baseline_tie}. }
\label{tab:core_results}

\resizebox{\textwidth}{!}{%
\setlength{\tabcolsep}{2pt}
\large
\begin{tabular}{l l ccccc c cccccc}
\toprule

& & \multicolumn{5}{c}{\textbf{Pairwise Accuracy (\%) $\uparrow$}} 
&
& \multicolumn{6}{c}{\textbf{Human Correlation $\uparrow$}} \\

\cmidrule(lr){3-7}
\cmidrule(lr){9-14}

&
& \textbf{Img+Txt}
& \textbf{Text}
& \multicolumn{2}{c}{\textbf{None}}
& \textbf{Avg.}
&
& \multicolumn{2}{c}{\textbf{VP2}}
& \multicolumn{2}{c}{\textbf{VF2}}
& \multicolumn{2}{c}{\textbf{Avg.}} \\

\cmidrule(lr){5-6}
\cmidrule(lr){9-10}
\cmidrule(lr){11-12}
\cmidrule(lr){13-14}

\textbf{Method}
& \textbf{Backbone}
& ImplB$^p$
& VP2$^p$
& PhyDEx$^p$
& VF2$^p$
&
&
& $\rho$
& $r$
& $\rho$
& $r$
& $\rho$
& $r$ \\

&
& {\scriptsize (R--G)}
& {\scriptsize (G--G)}
& {\scriptsize (R--G)}
& {\scriptsize (G--G)}
&
&
&
&
&
&
&
& \\

\midrule

$1$ VP2-AutoEval
& \small{VideoCon (7B)}
& 28.0
& 28.5
& 30.5
& 36.3
& 32.1
&& \textbf{0.359}
& \textbf{0.362}
& 0.235
& 0.240
& 0.298
& 0.302 \\

$2$ V-JEPA Surprise
& \small{V-JEPA2-H/16 (600M)}
& 33.3
& 46.3
& 49.0
& 45.5
& 46.6
&& 0.065
& 0.065
& 0.121
& 0.133
& 0.093
& 0.099 \\

$3$ VideoScore2
& \small{Qwen2.5-VL (7B)}
& 68.7
& 49.7
& 60.6
& 65.0
& 59.9
&& 0.197
& 0.160
& 0.462
& 0.430
& 0.336
& 0.301 \\

$4$ Gemini-3.1-Flash-Lite
& \small{N/A}
& 91.3
& 61.8
& 89.0
& 74.5
& 77.3
&& 0.302
& 0.303
& 0.419
& 0.409
& 0.362
& 0.357 \\

$5$ Gemini-3.1-Pro
& \small{N/A}
& 95.2
& \textbf{70.5}
& 86.2
& 73.7
& 78.1
&& 0.277
& 0.240
& 0.402
& 0.383
& 0.341
& 0.313 \\

\midrule

$6$ \textbf{\phyprobe{}}
& \small{PE-Core-G14-448 (2B)}
& \textbf{98.0}
& 66.1
& \textbf{98.9}
& \textbf{75.2}
& \textbf{82.4}
&& 0.293
& 0.302
& \textbf{0.596}
& \textbf{0.604}
& \textbf{0.458}
& \textbf{0.466} \\

\bottomrule
\end{tabular}%
}
\end{table*}
$\triangleright$ \textbf{\phyprobe{} is a strong physical consistency evaluator.}
\phyprobe{} achieves the highest pairwise accuracy on three out of four 
physical benchmarks as shown in Table~\ref{tab:core_results}, reaching 
98.0--98.9\% on R-G pairs and 66.1--75.2\% on harder G-G pairs. Unlike baselines, whose performance varies considerably across 
correspondence settings, \phyprobe{} remains strong with and without semantic correspondence.
It also achieves the strongest correlation with human judgments 
on VF2 (0.596/0.604, $\rho$/$r$) is competitive with VideoPhy2-AutoEval on VP2. Overall, \phyprobe{} delivers what a physical consistency evaluator 
should: bounded, consistently scaled scores that reliably order videos and reflect 
the severity of physical violations. This can also be seen in Figure~\ref{fig:ours_showcase} and~\ref{fig:ours_qual_extra} -- \phyprobe{} consistently 
aligns with physical plausibility, assigning low scores to 
physically consistent motion and high scores to clear violations 
such as ``object explosion'' or ``inconsistent dynamics'', even when human 
annotations are ambiguous or inconsistent. We feature more qualitative results in Figure~\ref{fig:ours_qual_i2v} and~\ref{fig:ours_qual_t2v} in Appendix~\ref{app:cases}. With the same V-JEPA2-H/16 encoder, \phyprobe{} outperforms V-JEPA Surprise by 13.9--48.1 points on all four benchmarks (Table~\ref{tab:ablation_backbone_invar}).

\begin{figure}[t]
    \centering

    \begin{subfigure}[t]{0.245\linewidth}
        \centering
        \includegraphics[width=\linewidth]{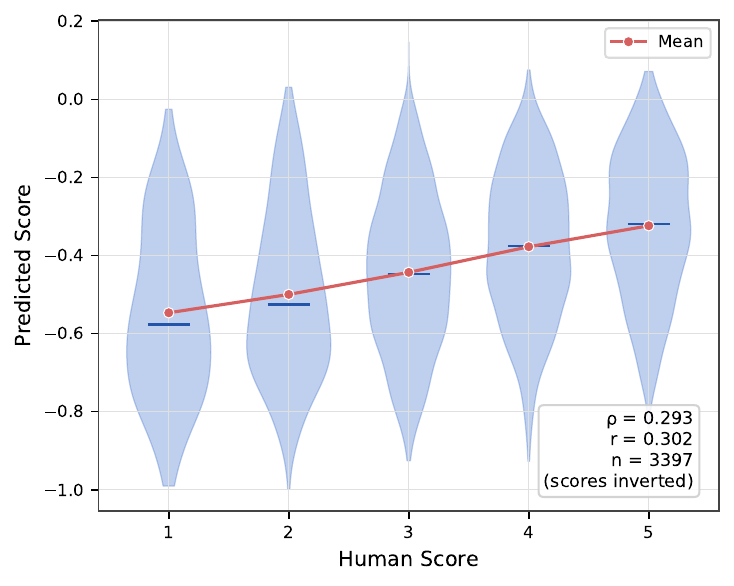}
        \caption{}
    \end{subfigure}
    \hfill
    \begin{subfigure}[t]{0.245\linewidth}
        \centering
        \includegraphics[width=\linewidth]{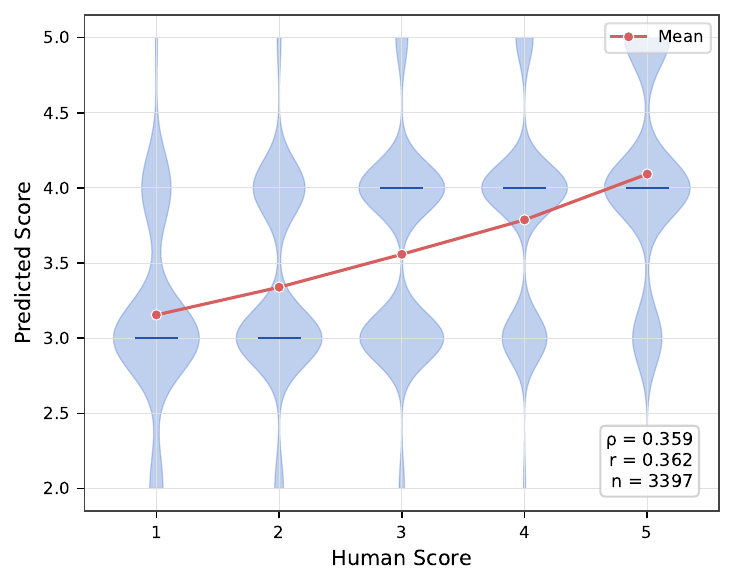}
        \caption{}
    \end{subfigure}
    \hfill
    \begin{subfigure}[t]{0.245\linewidth}
        \centering
        \includegraphics[width=\linewidth]{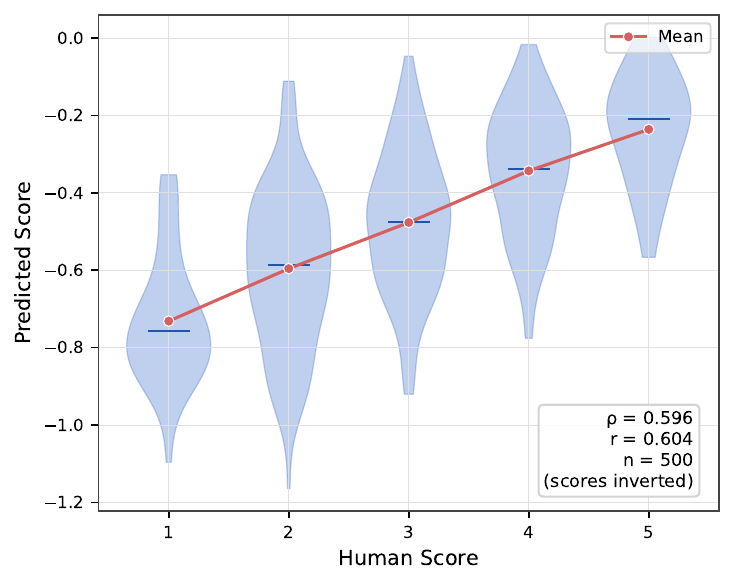}
        \caption{}
    \end{subfigure}
    \hfill
    \begin{subfigure}[t]{0.245\linewidth}
        \centering
        \includegraphics[width=\linewidth]{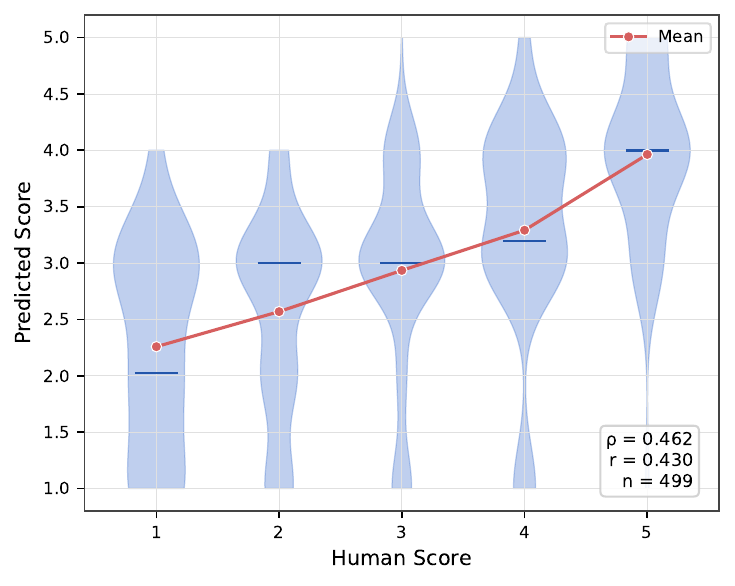}
        \caption{}
    \end{subfigure}

    \caption{
    Distribution of evaluator predictions conditioned on human rating levels on VideoPhy2 (VP2) and VideoFeedback2 (VF2). From left to right: (a) \phyprobe on VP2, (b) VP2-AutoEval on VP2, (c) \phyprobe on VF2, and (d) VideoScore2 on VF2. Each violin shows the distribution of predicted scores for videos assigned the same human rating, with the red line indicating the mean. For visualization, \phyprobe violation scores are inverted so that higher values correspond to greater physical plausibility, matching the direction of the baseline evaluators. Compared with prior methods, \phyprobe characterizes physical inconsistencies with greater granularity than the five-point human scale, providing a fluid assessment of error intensity beyond rigid categorical labels.
    }
    \label{fig:corr-vis}
\end{figure}

$\triangleright$ \textbf{\phyprobe{} balances alignment across annotation scales.}
From Table~\ref{tab:core_results}, we can see VP2-AutoEval (row 1) achieves the highest VP2 correlation 
($\rho\!=\!0.359$) but does not generalize well to VF2. Conversely, 
VideoScore2 (row 3) shows strong VF2 correlation 
($\rho\!=\!0.462$) but degrades on VP2 ($\rho\!=\!0.197$). 
Both methods appear calibrated to their respective annotation 
distributions rather than capturing transferable measurements of 
physical consistency. In contrast, \phyprobe{} achieves moderate correlation on VP2 ($\rho = 0.293$) and the strongest correlation on VF2 ($\rho = 0.596$) simultaneously, without collapsing on either. This balance reflects co-training across datasets and partial transfer between their supervision signals, although the leave-one-dataset-out analysis reveals meaningful dataset-specific dependence (Table~\ref{tab:lodo}). As further shown in Figure~\ref{fig:corr-vis}, prior 
methods produce predictions concentrated around discrete annotation 
levels, whereas \phyprobe{} yields more smoothly separated score 
distributions with broader intra-class variation. This behavior 
suggests that \phyprobe{} captures a continuous notion of violation 
severity beyond discrete human rating bins.

\begin{table*}[t]
\centering
\small
\setlength{\tabcolsep}{3.5pt}
\caption{\textbf{\phyprobe{} aligns with broader human preference beyond physical consistency.} 
Pairwise accuracy (\%) on human preference (quality) benchmarks~\citep{xu2026visionrewardfinegrainedmultidimensionalhuman, rapidata_hailuo02_marey_i2v_2025, rapidata_seedance1pro_i2v_2025, liu2025improvingvideogenerationhuman} under different tie-handling protocols. Despite being trained without explicit general-preference labels, \phyprobe{} achieves competitive performance across all datasets, categories (overall, MQ, VQ), and evaluation protocols. \phyprobe{} uses a model-side tie threshold at the 20th percentile.
}
\label{tab:quality_pairwise}
\resizebox{\textwidth}{!}{%
\begin{tabular}{l cc c cccc cc}
\toprule
\textbf{Method}
& \multicolumn{2}{c}{\textbf{MonetBench}} 
& \textbf{Rapidata-I2V} 
& \multicolumn{6}{c}{\textbf{VideoGen-RewardBench}} \\
\cmidrule(lr){2-3} \cmidrule(lr){5-10}
& w/ tie & w/o tie 
& w/o tie only 
& \multicolumn{2}{c}{Overall} 
& \multicolumn{2}{c}{MQ} 
& \multicolumn{2}{c}{VQ} \\
\cmidrule(lr){5-6} \cmidrule(lr){7-8} \cmidrule(lr){9-10}
& & & 
& w/ tie & w/o tie 
& w/ tie & w/o tie 
& w/ tie & w/o tie \\
\midrule
$1$ Random 
& 33.2 & 49.9 & 49.7 
& 33.2 & 49.7 
& 33.2 & 49.5 
& 33.3 & 49.8  \\
$2$ V-JEPA Surprise 
& 52.4 & 53.8 & 45.1 
& 45.2 & 44.2 
& 48.8 & 47.1 
& 48.7 & 47.6  \\
$3$ VideoScore2 
& 42.3 & 41.1 & 45.8 
& 57.4 & 58.9 
& 54.3 & 60.6  
& 55.3 & 60.1  \\
$4$ VideoPhy-2-AutoEval 
& 22.5 & 16.5 & 9.5 
& 24.6 & 19.5 
& 37.9 & 20.5 
& 34.4 & 20.4  \\
$5$ VisionReward 
& 68.2 & \textbf{72.2} & 47.1 
& 64.6 & 67.6  
& 54.6 & 61.3 
& 54.8 & 59.1  \\
$6$ VideoReward 
& 56.1 & 58.2 & 51.5 
& \textbf{69.7} & \textbf{73.6} 
& 60.3 & \textbf{75.2}  
& \textbf{63.6} & \textbf{75.9}  \\
\midrule
$7$ \textbf{\phyprobe{}} 
& \textbf{72.7} & 71.0 & \textbf{62.7} 
& 64.1 & 65.3 
& \textbf{62.3} & 65.9 
& 61.4 & 63.2 \\
\bottomrule
\end{tabular}%
}
\end{table*}

$\triangleright$ \textbf{\phyprobe{} transfers to general preference benchmarks without explicit preference-reward supervision.}
Although \phyprobe{} is trained on supervision indicative of physical consistency, without explicit general-preference labels, it achieves surprisingly strong performance on human 
preference benchmarks as shown in Table~\ref{tab:quality_pairwise}, 
outperforming dedicated reward models on MonetBench (72.7\% vs.\ 
VisionReward~\citep{xu2026visionrewardfinegrainedmultidimensionalhuman} 68.2\% and VideoReward~\citep{liu2025improvingvideogenerationhuman} 56.1\%, w/ tie), leading on Rapidata-I2V (62.7\%), and remaining competitive on RewardBench (65.3\% Overall w/o tie). We hypothesize this transfer occurs  because physics violations frequently co-occur with broader quality degradations in generated videos, making physical consistency a significant component of perceptual preference.

\subsection{Ablations}
\label{sec:results_ablations}


$\triangleright$ \textbf{Effect of \phyprobe{} loss components.} We ablate the effect of each loss term by removing it from \phyprobe{} training as shown in 
Table~\ref{tab:ablation_loss}.
Removing $\mathcal{L}_{\text{rank}}$ degrades pairwise ordering, 
particularly on VP2 and PhyDetEx. Without explicit preference supervision, 
the model lacks direct signal for relative comparison between 
videos of similar quality.
Removing $\mathcal{L}_{\text{mag}}$ leaves pairwise accuracy largely intact but causes the score range to explode: without 
regression targets, \phyprobe{} learns to order videos correctly but 
has no incentive to map violation scores to a bounded, interpretable scale.
Removing $\mathcal{L}_{\text{cal}}$ also causing score range to explode and collapses real--generated pairwise accuracy: without anchor points defining what constitutes 
a clearly consistent or clearly violated video, the model cannot 
assign appropriate scores to obvious cases, even though it may 
still rank ambiguous pairs well.
The full objective provides the best trade-off: trades some correlation for bounded range and strong pairwise 
accuracy, and alignment with human ratings.

\begin{table*}[!h]
\centering
\small
\setlength{\tabcolsep}{3.5pt}
\caption{
\textbf{Importance of individual loss components in \phyprobe{}.}
We compare \phyprobe{} with different loss components using the same backbone (PE-Core-G14-448).
The scoring range reflects the effective output range of each variant.
}
\label{tab:ablation_loss}
\begin{tabular}{l c c c c c cc cc c}
\toprule
\textbf{Variant}
& \multicolumn{4}{c}{\textbf{Pairwise Accuracy (\%)}} 
& \multicolumn{4}{c}{\textbf{Human Correlation}} 
& \textbf{Score} \\
\cmidrule(lr){2-5} \cmidrule(lr){6-9}
& \multicolumn{2}{c}{\textbf{None}} 
& \textbf{Txt} 
& \textbf{Img+Txt}
& \multicolumn{2}{c}{\textbf{VP2}} 
& \multicolumn{2}{c}{\textbf{VF2}} 
& \textbf{Range} \\
\cmidrule(lr){2-3} \cmidrule(lr){4-4} \cmidrule(lr){5-5} \cmidrule(lr){6-7} \cmidrule(lr){8-9}
& \scriptsize PhyDEx$^p$ & \scriptsize VF2$^p$
& \scriptsize VP2$^p$
& \scriptsize ImplB$^p$
& $\rho$ & $r$
& $\rho$ & $r$ 
& \\
& \scriptsize (R--G) & \scriptsize (G--G)
& \scriptsize (G--G)
& \scriptsize (R--G)
& & & & & \\
\midrule

\textbf{\phyprobe{} ($- \mathcal{L}_{\text{rank}}$)} 
& 94.7 & 75.6
& 61.9 & 99.2
& 0.311 & 0.316 
& 0.614 & 0.609 
& $[-0.07, 0.88]$ \\

\textbf{\phyprobe{} ($- \mathcal{L}_{\text{mag}}$)} 
& 98.7 & 75.3
& 62.3 & 94.1
& 0.315 & 0.316 
& 0.584 & 0.583 
& $[-1.89, 3.45]$ \\

\textbf{\phyprobe{} ($- \mathcal{L}_{\text{cal}}$)} 
& 68.7 & \textbf{79.3} 
& 64.4 & 84.9 
& 0.363 & 0.364 
& 0.664 & 0.656 
& $[-0.40, 4.13]$ \\
\midrule
\textbf{\phyprobe{}} 
& \textbf{98.9} & 75.2
& \textbf{66.1} & \textbf{98.0}
& 0.293 & 0.302 
& 0.596 & 0.604 
& $[-0.00,1.15]$ \\

\bottomrule
\end{tabular}
\end{table*}

\begin{table*}[t]
\centering
\small
\caption{
\textbf{Effect of backbone representations on \phyprobe{}}.
We train and evaluate \phyprobe{} with different pretrained video encoders, reporting 
pairwise accuracy (\%) and correlation with human judgments.
Pairwise accuracy is decomposed by correspondence type (image+text, text, none) 
and pair type (real (R) vs generated (G)).
Benchmark$^p$ indicates the benchmark is reordered in pairwise manner without ties.
Correlation metrics (Spearman $\rho$, Pearson $r$) measure alignment with human 
judgments on VideoPhy2 (VP2) and VideoFeedback2 (VF2).}
\label{tab:ablation_backbone}
\resizebox{0.9\textwidth}{!}{%
\setlength{\tabcolsep}{2pt}
\large
\begin{tabular}{l c cccc c cccc}
\toprule
& & \multicolumn{4}{c}{\textbf{Pairwise Accuracy (\%) $\uparrow$}} 
&
& \multicolumn{4}{c}{\textbf{Human Correlation $\uparrow$}} \\
\cmidrule(lr){3-6} \cmidrule(lr){8-11}
& & \textbf{Img+Txt} & \textbf{Text} 
& \multicolumn{2}{c}{\textbf{None}} 
&
& \multicolumn{2}{c}{\textbf{VP2}} 
& \multicolumn{2}{c}{\textbf{VF2}} \\
\cmidrule(lr){5-6} \cmidrule(lr){8-9} \cmidrule(lr){10-11}
\textbf{\phyprobe{}} & \textbf{Params}
& ImplB$^p$ & VP2$^p$ & PhyDEx$^p$ & VF2$^p$
&
& $\rho$ & $r$
& $\rho$ & $r$ \\
\textbf{Backbone} &
& {\scriptsize (R--G)} & {\scriptsize (G--G)}
& {\scriptsize (R--G)} & {\scriptsize (G--G)}
& & & & & \\
\midrule
$1$ Wan2.2 VAE & 300M 
& 71.7 & 58.3 & 67.1 & 74.3
&& 0.186 & 0.194 & 0.562 & 0.581 \\
$2$ V-JEPA2-VIT-L/16 & 300M 
& 76.1 & 60.6 & 88.8 & 74.2
&& 0.228 & 0.206 & 0.564 & 0.580 \\
$3$ V-JEPA2-VIT-H/16 & 600M 
& 81.4 & 60.2 & 70.7 & 76.3
&& 0.301 & 0.303 & 0.623 & 0.630 \\
$4$ V-JEPA2-VIT-G/16 & 1B 
& 88.6 & 60.9 & 93.0 & 72.6
&& 0.279 & 0.288 & 0.566 & 0.580 \\
$5$ PE-Spatial-G14-448 & 2B 
& 98.7 & 61.3 & 99.5 & 73.0
&& 0.244 & 0.253 & 0.566 & 0.573 \\
$6$ PE-Core-G14-448 & 2B 
& 98.0 & 66.1 & 98.9 & 75.2
&& 0.293 & 0.302 & 0.596 & 0.604 \\
\bottomrule
\end{tabular}%
}
\end{table*}

$\triangleright$ \textbf{Effect of Backbone Representations.}
We evaluate \phyprobe{} with different frozen pretrained video encoders, as shown in Table~\ref{tab:ablation_backbone}. We compare state-of-the-art video tokenizers (Wan-2.2 VAE~\citep{wan22-vae}), latent world model architectures (V-JEPA2~\citep{assran2025vjepa2selfsupervisedvideo}), and weakly supervised pretrained visual encoders (Perception Encoder~\citep{bolya2025perceptionencoderbestvisual}). We observe substantial performance differences across representation types, indicating that physical consistency evaluation is highly dependent on the quality of the underlying spatiotemporal representation space. In particular, weakly supervised pretrained visual encoders (PE) achieve the strongest and most consistent performance across both pairwise evaluation and human correlation benchmarks. We hypothesize that their large-scale pretraining and robust visual representation learning pipeline better preserve object structure, motion continuity, and interaction dynamics. V-JEPA2 also performs competitively, suggesting that predictive spatiotemporal representations are naturally aligned with modeling physical dynamics, whereas latent representations optimized primarily for reconstruction exhibit weaker transfer.


$\triangleright$ \textbf{Effect of Spatiotemporal Aggregation.} We compare different spatiotemporal aggregation strategies for mapping frame-level features to a scalar score in Table~\ref{tab:ablation_head}. Within each frame, patch tokens are spatially aggregated via mean pooling to produce a single frame-level descriptor. Across frames, a statistics-based aggregation (mean, standard deviation, and max pooling concatenated) provides a strong performance-efficiency trade-off. Despite its simplicity, this approach captures key spatiotemporal variations indicative of physical inconsistencies, while avoiding the optimization challenges and computational overhead of attention-based aggregation. We therefore used statistical aggregation head (mean+std+max) in our final model, as it achieves the best trade-off between accuracy and model complexity.

\begin{table}[!h]
\centering
\small
\setlength{\tabcolsep}{4.5pt}
\caption{
Ablation of projection head designs. 
We compare different spatiotemporal aggregation strategies in terms of performance and efficiency, under the same PECore backbone.
Rel. denotes relative trainable parameter count normalized to the Mean (MLP) head.}
\label{tab:ablation_head}
\begin{tabular}{lccc cc cc}
\toprule
\textbf{Head} 
& \textbf{Trainable} 
& \textbf{Input} 
& \textbf{Rel. (×)} 
& \multicolumn{2}{c}{\textbf{VideoPhy2}} 
& \multicolumn{2}{c}{\textbf{VideoFeedback2}} \\
\cmidrule(lr){5-6} \cmidrule(lr){7-8}
& \textbf{Params} & \textbf{Dim} & 
& Acc & $\rho$ 
& Acc & $\rho$ \\
\midrule

Mean (MLP) 
& 0.36M 
& $[B, D]$ 
& 1$\times$ 
& 61.0 & 0.277 
& 73.2 & 0.578 \\

Attention Pooling 
& 6.92M 
& $[B, T, D]$ 
& 19.2$\times$ 
& 63.7 & 0.292 
& 75.0 & 0.612 \\



\textbf{Stat (mean+std+max)}
& 1.02M 
& $[B, 3D]$ 
& 2.8$\times$ 
& 66.1 & 0.293 
& 75.2 & 0.596 \\

\bottomrule
\end{tabular}
\end{table}

$\triangleright$ \textbf{Further ablations (Appendix~A.2).} \phyprobe{}'s advantage is robust to the minimum score gap used to construct evaluation pairs (Table~\ref{tab:videophy2_gap_threshold}); its learned signal transfers partially across supervision sources while retaining dataset-specific dependence (Table~\ref{tab:lodo}); and violation-flooring normalization improves pairwise accuracy over linear normalization on all four benchmarks (Table~\ref{tab:linear_mapping}).
\section{Limitations}
\label{sec:limiations}

Current spatiotemporal representations model long-range causal dependencies imperfectly, so violations that emerge only over extended horizons may be under-encoded~\citep{xue2025seeingarrowtimelarge}; conversely, very brief or instantaneous violations (e.g., abrupt state changes) may not persist long enough to register in the aggregated representation, and unusual but physically valid motions in real videos can receive elevated scores. Failure cases are shown in Figure~\ref{fig:ours_qual_fail} (Appendix~\ref{app:cases}). \phyprobe{} is also trained to approximate human-perceived plausibility, the operational definition used by existing benchmarks; since those judgments correlate with other perceptual factors such as overall quality, our model may partially capture them as well. A controlled evaluation varying physical consistency while holding quality, scene, prompt, and generator fixed would give stronger evidence of disentanglement, but to our knowledge no benchmark with reliable human annotations enables this, a limitation shared by current physical-consistency benchmarks and evaluators.

\section{Conclusion}
\label{sec:conclusion}

We introduced \phyprobe{}, 
an evaluator for physical consistency in generated videos. 
Through a unified training objective combining pairwise ranking, regression on noisy scalar annotations, and anchor-based calibration across heterogeneous data sources, \phyprobe{} produces continuous, stably anchored violation scores that capture the severity of physical inconsistencies.
Empirically, \phyprobe{} achieves strong performance across diverse evaluation settings, maintaining robust pairwise ordering even in challenging regimes where prior approaches degrade, while also aligning well with human judgments. 
Notably, \phyprobe{} generalizes to broader human-preference benchmarks despite being trained on supervision selected to indicate physical consistency rather than general preference.
We hope this work encourages future research towards 
the development of more reliable and scalable metrics for assessing the physical fidelity of video generation systems.



\clearpage
\bibliographystyle{abbrvnat}
\bibliography{main}

@misc{bear2022physionevaluatingphysicalprediction,
      title={Physion: Evaluating Physical Prediction from Vision in Humans and Machines}, 
      author={Daniel M. Bear and Elias Wang and Damian Mrowca and Felix J. Binder and Hsiao-Yu Fish Tung and R. T. Pramod and Cameron Holdaway and Sirui Tao and Kevin Smith and Fan-Yun Sun and Li Fei-Fei and Nancy Kanwisher and Joshua B. Tenenbaum and Daniel L. K. Yamins and Judith E. Fan},
      year={2022},
      eprint={2106.08261},
      archivePrefix={arXiv},
      primaryClass={cs.AI},
      url={https://arxiv.org/abs/2106.08261}, 
}

@misc{lin2025brokenvideosbenchmarkdatasetfinegrained,
      title={BrokenVideos: A Benchmark Dataset for Fine-Grained Artifact Localization in AI-Generated Videos}, 
      author={Jiahao Lin and Weixuan Peng and Bojia Zi and Yifeng Gao and Xianbiao Qi and Xingjun Ma and Yu-Gang Jiang},
      year={2025},
      eprint={2506.20103},
      archivePrefix={arXiv},
      primaryClass={cs.CV},
      url={https://arxiv.org/abs/2506.20103}, 
}

@misc{bolya2025perceptionencoderbestvisual,
      title={Perception Encoder: The best visual embeddings are not at the output of the network}, 
      author={Daniel Bolya and Po-Yao Huang and Peize Sun and Jang Hyun Cho and Andrea Madotto and Chen Wei and Tengyu Ma and Jiale Zhi and Jathushan Rajasegaran and Hanoona Rasheed and Junke Wang and Marco Monteiro and Hu Xu and Shiyu Dong and Nikhila Ravi and Daniel Li and Piotr Dollár and Christoph Feichtenhofer},
      year={2025},
      eprint={2504.13181},
      archivePrefix={arXiv},
      primaryClass={cs.CV},
      url={https://arxiv.org/abs/2504.13181}, 
}

@misc{gao2025davidxr1detectingaigeneratedvideos,
      title={DAVID-XR1: Detecting AI-Generated Videos with Explainable Reasoning}, 
      author={Yifeng Gao and Yifan Ding and Hongyu Su and Juncheng Li and Yunhan Zhao and Lin Luo and Zixing Chen and Li Wang and Xin Wang and Yixu Wang and Xingjun Ma and Yu-Gang Jiang},
      year={2025},
      eprint={2506.14827},
      archivePrefix={arXiv},
      primaryClass={cs.CV},
      url={https://arxiv.org/abs/2506.14827}, 
}

@misc{bai2025impossiblevideos,
      title={Impossible Videos}, 
      author={Zechen Bai and Hai Ci and Mike Zheng Shou},
      year={2025},
      eprint={2503.14378},
      archivePrefix={arXiv},
      primaryClass={cs.CV},
      url={https://arxiv.org/abs/2503.14378}, 
}

@misc{riochet2020intphysframeworkbenchmarkvisual,
      title={IntPhys: A Framework and Benchmark for Visual Intuitive Physics Reasoning}, 
      author={Ronan Riochet and Mario Ynocente Castro and Mathieu Bernard and Adam Lerer and Rob Fergus and Véronique Izard and Emmanuel Dupoux},
      year={2020},
      eprint={1803.07616},
      archivePrefix={arXiv},
      primaryClass={cs.AI},
      url={https://arxiv.org/abs/1803.07616}, 
}

@misc{bordes2025intphys2benchmarkingintuitive,
      title={IntPhys 2: Benchmarking Intuitive Physics Understanding In Complex Synthetic Environments}, 
      author={Florian Bordes and Quentin Garrido and Justine T Kao and Adina Williams and Michael Rabbat and Emmanuel Dupoux},
      year={2025},
      eprint={2506.09849},
      archivePrefix={arXiv},
      primaryClass={cs.CV},
      url={https://arxiv.org/abs/2506.09849}, 
}

@misc{li2025worldmodelbenchjudgingvideogeneration,
      title={WorldModelBench: Judging Video Generation Models As World Models}, 
      author={Dacheng Li and Yunhao Fang and Yukang Chen and Shuo Yang and Shiyi Cao and Justin Wong and Michael Luo and Xiaolong Wang and Hongxu Yin and Joseph E. Gonzalez and Ion Stoica and Song Han and Yao Lu},
      year={2025},
      eprint={2502.20694},
      archivePrefix={arXiv},
      primaryClass={cs.CV},
      url={https://arxiv.org/abs/2502.20694}, 
}

@misc{bansal2024videophyevaluatingphysicalcommonsense,
      title={VideoPhy: Evaluating Physical Commonsense for Video Generation}, 
      author={Hritik Bansal and Zongyu Lin and Tianyi Xie and Zeshun Zong and Michal Yarom and Yonatan Bitton and Chenfanfu Jiang and Yizhou Sun and Kai-Wei Chang and Aditya Grover},
      year={2024},
      eprint={2406.03520},
      archivePrefix={arXiv},
      primaryClass={cs.CV},
      url={https://arxiv.org/abs/2406.03520}, 
}

@misc{bansal2025videophy2challengingactioncentricphysical,
      title={VideoPhy-2: A Challenging Action-Centric Physical Commonsense Evaluation in Video Generation}, 
      author={Hritik Bansal and Clark Peng and Yonatan Bitton and Roman Goldenberg and Aditya Grover and Kai-Wei Chang},
      year={2025},
      eprint={2503.06800},
      archivePrefix={arXiv},
      primaryClass={cs.CV},
      url={https://arxiv.org/abs/2503.06800}, 
}

@misc{he2024videoscorebuildingautomaticmetrics,
      title={VideoScore: Building Automatic Metrics to Simulate Fine-grained Human Feedback for Video Generation}, 
      author={Xuan He and Dongfu Jiang and Ge Zhang and Max Ku and Achint Soni and Sherman Siu and Haonan Chen and Abhranil Chandra and Ziyan Jiang and Aaran Arulraj and Kai Wang and Quy Duc Do and Yuansheng Ni and Bohan Lyu and Yaswanth Narsupalli and Rongqi Fan and Zhiheng Lyu and Yuchen Lin and Wenhu Chen},
      year={2024},
      eprint={2406.15252},
      archivePrefix={arXiv},
      primaryClass={cs.CV},
      url={https://arxiv.org/abs/2406.15252}, 
}

@misc{he2025videoscore2thinkscoregenerative,
      title={VideoScore2: Think before You Score in Generative Video Evaluation}, 
      author={Xuan He and Dongfu Jiang and Ping Nie and Minghao Liu and Zhengxuan Jiang and Mingyi Su and Wentao Ma and Junru Lin and Chun Ye and Yi Lu and Keming Wu and Benjamin Schneider and Quy Duc Do and Zhuofeng Li and Yiming Jia and Yuxuan Zhang and Guo Cheng and Haozhe Wang and Wangchunshu Zhou and Qunshu Lin and Yuanxing Zhang and Ge Zhang and Wenhao Huang and Wenhu Chen},
      year={2025},
      eprint={2509.22799},
      archivePrefix={arXiv},
      primaryClass={cs.CV},
      url={https://arxiv.org/abs/2509.22799}, 
}

@misc{motamed2025generativevideomodelsunderstand_physicsiq,
      title={Do generative video models understand physical principles?}, 
      author={Saman Motamed and Laura Culp and Kevin Swersky and Priyank Jaini and Robert Geirhos},
      year={2025},
      eprint={2501.09038},
      archivePrefix={arXiv},
      primaryClass={cs.CV},
      url={https://arxiv.org/abs/2501.09038}, 
}

@misc{meng2024worldsimulatorcraftingphysical,
      title={Towards World Simulator: Crafting Physical Commonsense-Based Benchmark for Video Generation}, 
      author={Fanqing Meng and Jiaqi Liao and Xinyu Tan and Wenqi Shao and Quanfeng Lu and Kaipeng Zhang and Yu Cheng and Dianqi Li and Yu Qiao and Ping Luo},
      year={2024},
      eprint={2410.05363},
      archivePrefix={arXiv},
      primaryClass={cs.CV},
      url={https://arxiv.org/abs/2410.05363}, 
}

@misc{nvidia2025cosmosworldfoundationmodel,
      title={Cosmos World Foundation Model Platform for Physical AI}, 
      author={NVIDIA and : and Niket Agarwal and Arslan Ali and Maciej Bala and Yogesh Balaji and Erik Barker and Tiffany Cai and Prithvijit Chattopadhyay and Yongxin Chen and Yin Cui and Yifan Ding and Daniel Dworakowski and Jiaojiao Fan and Michele Fenzi and Francesco Ferroni and Sanja Fidler and Dieter Fox and Songwei Ge and Yunhao Ge and Jinwei Gu and Siddharth Gururani and Ethan He and Jiahui Huang and Jacob Huffman and Pooya Jannaty and Jingyi Jin and Seung Wook Kim and Gergely Klár and Grace Lam and Shiyi Lan and Laura Leal-Taixe and Anqi Li and Zhaoshuo Li and Chen-Hsuan Lin and Tsung-Yi Lin and Huan Ling and Ming-Yu Liu and Xian Liu and Alice Luo and Qianli Ma and Hanzi Mao and Kaichun Mo and Arsalan Mousavian and Seungjun Nah and Sriharsha Niverty and David Page and Despoina Paschalidou and Zeeshan Patel and Lindsey Pavao and Morteza Ramezanali and Fitsum Reda and Xiaowei Ren and Vasanth Rao Naik Sabavat and Ed Schmerling and Stella Shi and Bartosz Stefaniak and Shitao Tang and Lyne Tchapmi and Przemek Tredak and Wei-Cheng Tseng and Jibin Varghese and Hao Wang and Haoxiang Wang and Heng Wang and Ting-Chun Wang and Fangyin Wei and Xinyue Wei and Jay Zhangjie Wu and Jiashu Xu and Wei Yang and Lin Yen-Chen and Xiaohui Zeng and Yu Zeng and Jing Zhang and Qinsheng Zhang and Yuxuan Zhang and Qingqing Zhao and Artur Zolkowski},
      year={2025},
      eprint={2501.03575},
      archivePrefix={arXiv},
      primaryClass={cs.CV},
      url={https://arxiv.org/abs/2501.03575}, 
}

@misc{yang2025cogvideoxtexttovideodiffusionmodels,
      title={CogVideoX: Text-to-Video Diffusion Models with An Expert Transformer}, 
      author={Zhuoyi Yang and Jiayan Teng and Wendi Zheng and Ming Ding and Shiyu Huang and Jiazheng Xu and Yuanming Yang and Wenyi Hong and Xiaohan Zhang and Guanyu Feng and Da Yin and Yuxuan Zhang and Weihan Wang and Yean Cheng and Bin Xu and Xiaotao Gu and Yuxiao Dong and Jie Tang},
      year={2025},
      eprint={2408.06072},
      archivePrefix={arXiv},
      primaryClass={cs.CV},
      url={https://arxiv.org/abs/2408.06072}, 
}

@misc{wan2025wanopenadvancedlargescale,
      title={Wan: Open and Advanced Large-Scale Video Generative Models}, 
      author={Team Wan and Ang Wang and Baole Ai and Bin Wen and Chaojie Mao and Chen-Wei Xie and Di Chen and Feiwu Yu and Haiming Zhao and Jianxiao Yang and Jianyuan Zeng and Jiayu Wang and Jingfeng Zhang and Jingren Zhou and Jinkai Wang and Jixuan Chen and Kai Zhu and Kang Zhao and Keyu Yan and Lianghua Huang and Mengyang Feng and Ningyi Zhang and Pandeng Li and Pingyu Wu and Ruihang Chu and Ruili Feng and Shiwei Zhang and Siyang Sun and Tao Fang and Tianxing Wang and Tianyi Gui and Tingyu Weng and Tong Shen and Wei Lin and Wei Wang and Wei Wang and Wenmeng Zhou and Wente Wang and Wenting Shen and Wenyuan Yu and Xianzhong Shi and Xiaoming Huang and Xin Xu and Yan Kou and Yangyu Lv and Yifei Li and Yijing Liu and Yiming Wang and Yingya Zhang and Yitong Huang and Yong Li and You Wu and Yu Liu and Yulin Pan and Yun Zheng and Yuntao Hong and Yupeng Shi and Yutong Feng and Zeyinzi Jiang and Zhen Han and Zhi-Fan Wu and Ziyu Liu},
      year={2025},
      eprint={2503.20314},
      archivePrefix={arXiv},
      primaryClass={cs.CV},
      url={https://arxiv.org/abs/2503.20314}, 
}

@article{article,
author = {Huynh-Thu, Q. and Ghanbari, Mohammed},
year = {2008},
month = {02},
pages = {800 - 801},
title = {Scope of validity of PSNR in image/video quality assessment},
volume = {44},
journal = {Electronics Letters},
doi = {10.1049/el:20080522}
}

@misc{google_gemini3,
  title        = {Gemini 3 Developer Guide},
  author       = {{Google AI for Developers}},
  year         = {2026},
  howpublished = {\url{https://ai.google.dev/gemini-api/docs/gemini-3}},
  note         = {Accessed: 2026-01-15}
}

@article{WeihsEtAl2022InfLevel,
  title={Benchmarking Progress to Infant-Level Physical Reasoning in AI},
  author={Luca Weihs and Amanda Rose Yuile and Ren\'{e}e Baillargeon and Cynthia Fisher and Gary Marcus and Roozbeh Mottaghi and Aniruddha Kembhavi},
  journal={TMLR},
  year={2022}
}

@article{BAILLARGEON1985191,
title = {Object permanence in five-month-old infants},
journal = {Cognition},
volume = {20},
number = {3},
pages = {191-208},
year = {1985},
issn = {0010-0277},
doi = {https://doi.org/10.1016/0010-0277(85)90008-3},
url = {https://www.sciencedirect.com/science/article/pii/0010027785900083},
author = {Renée Baillargeon and Elizabeth S. Spelke and Stanley Wasserman}
}

@article{margoni2024violation,
  author = {Margoni, Francesco and Surian, Luca and Baillargeon, Ren\'{e}e},
  title = {The violation-of-expectation paradigm: A conceptual overview},
  journal = {Psychological Review},
  year = {2024},
  volume = {131},
  number = {3},
  pages = {716--748},
  doi = {10.1037/rev0000450}
}

@misc{motamed2025travlrecipemakingvideolanguage,
      title={TRAVL: A Recipe for Making Video-Language Models Better Judges of Physics Implausibility}, 
      author={Saman Motamed and Minghao Chen and Luc Van Gool and Iro Laina},
      year={2025},
      eprint={2510.07550},
      archivePrefix={arXiv},
      primaryClass={cs.CV},
      url={https://arxiv.org/abs/2510.07550}, 
}

@misc{assran2025vjepa2selfsupervisedvideo,
      title={V-JEPA 2: Self-Supervised Video Models Enable Understanding, Prediction and Planning}, 
      author={Mido Assran and Adrien Bardes and David Fan and Quentin Garrido and Russell Howes and Mojtaba and Komeili and Matthew Muckley and Ammar Rizvi and Claire Roberts and Koustuv Sinha and Artem Zholus and Sergio Arnaud and Abha Gejji and Ada Martin and Francois Robert Hogan and Daniel Dugas and Piotr Bojanowski and Vasil Khalidov and Patrick Labatut and Francisco Massa and Marc Szafraniec and Kapil Krishnakumar and Yong Li and Xiaodong Ma and Sarath Chandar and Franziska Meier and Yann LeCun and Michael Rabbat and Nicolas Ballas},
      year={2025},
      eprint={2506.09985},
      archivePrefix={arXiv},
      primaryClass={cs.AI},
      url={https://arxiv.org/abs/2506.09985}, 
}

@article{nightingale2019detect,
  title     = {Can people detect errors in shadows and reflections?},
  author    = {Nightingale, Sophie J. and Wade, Kimberley A. and Farid, Hany and Watson, Derrick G.},
  journal   = {Attention, Perception, \& Psychophysics},
  volume    = {81},
  number    = {8},
  pages     = {2917--2943},
  year      = {2019},
  month     = {Nov},
  doi       = {10.3758/s13414-019-01773-w},
  pmid      = {31254262},
  pmcid     = {PMC6856028}
}

@article{nightingale2017identify,
  title     = {Can people identify original and manipulated photos of real-world scenes?},
  author    = {Nightingale, Sophie J. and Wade, Kimberley A. and Watson, Derrick G.},
  journal   = {Cognitive Research: Principles and Implications},
  volume    = {2},
  number    = {1},
  pages     = {30},
  year      = {2017},
  month     = {Jul},
  doi       = {10.1186/s41235-017-0067-2},
  pmid      = {28776002},
  pmcid     = {PMC5514174}
}

@article{nightingale2022ai,
  title     = {AI-synthesized faces are indistinguishable from real faces and more trustworthy},
  author    = {Nightingale, Sophie J. and Farid, Hany},
  journal   = {Proceedings of the National Academy of Sciences},
  volume    = {119},
  number    = {8},
  pages     = {e2120481119},
  year      = {2022},
  month     = {Feb},
  doi       = {10.1073/pnas.2120481119},
  pmid      = {35165187},
  pmcid     = {PMC8872790}
}

@article{voetrigger2022,
title = {Violations of expectation trigger infants to search for explanations},
journal = {Cognition},
volume = {218},
pages = {104942},
year = {2022},
issn = {0010-0277},
doi = {https://doi.org/10.1016/j.cognition.2021.104942},
url = {https://www.sciencedirect.com/science/article/pii/S0010027721003656},
author = {Jasmin Perez and Lisa Feigenson}
}

@article{elecphy2024,
    author = {Balaban, Halely and Smith, Kevin A. and Tenenbaum, Joshua B. and Ullman, Tomer D.},
    title = {Electrophysiology Reveals That Intuitive Physics Guides Visual Tracking and Working Memory},
    journal = {Open Mind},
    volume = {8},
    pages = {1425-1446},
    year = {2024},
    month = {11},
    issn = {2470-2986},
    doi = {10.1162/opmi_a_00174},
    url = {https://doi.org/10.1162/opmi_a_00174},
    eprint = {https://direct.mit.edu/opmi/article-pdf/doi/10.1162/opmi_a_00174/2483508/opmi_a_00174.pdf},
}

@article{Huber1964RobustEO,
  title={Robust Estimation of a Location Parameter},
  author={Peter J. Huber},
  journal={Annals of Mathematical Statistics},
  year={1964},
  volume={35},
  pages={492-518},
  url={https://api.semanticscholar.org/CorpusID:121252793}
}

@article{Bradley1952RankAO,
  title={Rank Analysis of Incomplete Block Designs: I. The Method of Paired Comparisons},
  author={Ralph Allan Bradley and Milton E. Terry},
  journal={Biometrika},
  year={1952},
  volume={39},
  pages={324},
  url={https://api.semanticscholar.org/CorpusID:125209808}
}

@misc{kay2017kineticshumanactionvideo,
      title={The Kinetics Human Action Video Dataset}, 
      author={Will Kay and Joao Carreira and Karen Simonyan and Brian Zhang and Chloe Hillier and Sudheendra Vijayanarasimhan and Fabio Viola and Tim Green and Trevor Back and Paul Natsev and Mustafa Suleyman and Andrew Zisserman},
      year={2017},
      eprint={1705.06950},
      archivePrefix={arXiv},
      primaryClass={cs.CV},
      url={https://arxiv.org/abs/1705.06950}, 
}

@misc{carreira2022shortnotekinetics700human,
      title={A Short Note on the Kinetics-700 Human Action Dataset}, 
      author={Joao Carreira and Eric Noland and Chloe Hillier and Andrew Zisserman},
      year={2022},
      eprint={1907.06987},
      archivePrefix={arXiv},
      primaryClass={cs.CV},
      url={https://arxiv.org/abs/1907.06987}, 
}

@misc{zhou2025paibenchcomprehensivebenchmarkphysical,
      title={PAI-Bench: A Comprehensive Benchmark For Physical AI}, 
      author={Fengzhe Zhou and Jiannan Huang and Jialuo Li and Deva Ramanan and Humphrey Shi},
      year={2025},
      eprint={2512.01989},
      archivePrefix={arXiv},
      primaryClass={cs.CV},
      url={https://arxiv.org/abs/2512.01989}, 
}

@misc{wang2025phydetexdetectingexplainingphysical,
      title={PhyDetEx: Detecting and Explaining the Physical Plausibility of T2V Models}, 
      author={Zeqing Wang and Keze Wang and Lei Zhang},
      year={2025},
      eprint={2512.01843},
      archivePrefix={arXiv},
      primaryClass={cs.CV},
      url={https://arxiv.org/abs/2512.01843}, 
}

@misc{liu2025improvingvideogenerationhuman,
      title={Improving Video Generation with Human Feedback}, 
      author={Jie Liu and Gongye Liu and Jiajun Liang and Ziyang Yuan and Xiaokun Liu and Mingwu Zheng and Xiele Wu and Qiulin Wang and Menghan Xia and Xintao Wang and Xiaohong Liu and Fei Yang and Pengfei Wan and Di Zhang and Kun Gai and Yujiu Yang and Wanli Ouyang},
      year={2025},
      eprint={2501.13918},
      archivePrefix={arXiv},
      primaryClass={cs.CV},
      url={https://arxiv.org/abs/2501.13918}, 
}

@misc{xu2026visionrewardfinegrainedmultidimensionalhuman,
      title={VisionReward: Fine-Grained Multi-Dimensional Human Preference Learning for Image and Video Generation}, 
      author={Jiazheng Xu and Yu Huang and Jiale Cheng and Yuanming Yang and Jiajun Xu and Yuan Wang and Wenbo Duan and Shen Yang and Qunlin Jin and Shurun Li and Jiayan Teng and Zhuoyi Yang and Wendi Zheng and Xiao Liu and Dan Zhang and Ming Ding and Xiaohan Zhang and Xiaotao Gu and Shiyu Huang and Minlie Huang and Jie Tang and Yuxiao Dong},
      year={2026},
      eprint={2412.21059},
      archivePrefix={arXiv},
      primaryClass={cs.CV},
      url={https://arxiv.org/abs/2412.21059}, 
}

@misc{rapidata_hailuo02_marey_i2v_2025,
  author       = {Rapidata},
  title        = {Image-to-Video Human Preference: Hailuo-02 vs Marey},
  year         = {2025},
  howpublished = {\url{https://huggingface.co/datasets/Rapidata/image-to-video-human-preference-hailuo-02-marey}},
  note         = {~6k human preference annotations from ~2k annotators on image-to-video generation},
}

@misc{rapidata_seedance1pro_i2v_2025,
  author       = {Rapidata},
  title        = {Image-to-Video Human Preference: Seedance-1-Pro Benchmark},
  year         = {2025},
  howpublished = {\url{https://huggingface.co/datasets/Rapidata/image-to-video-human-preference-seedance-1-pro}},
  note         = {Human preference dataset with pairwise video comparisons for evaluating image-to-video models},
}

@misc{seedance2026seedance20advancingvideo,
      title={Seedance 2.0: Advancing Video Generation for World Complexity}, 
      author={Team Seedance and De Chen and Liyang Chen and Xin Chen and Ying Chen and Zhuo Chen and Zhuowei Chen and Feng Cheng and Tianheng Cheng and Yufeng Cheng and Mojie Chi and Xuyan Chi and Jian Cong and Qinpeng Cui and Fei Ding and Qide Dong and Yujiao Du and Haojie Duanmu and Junliang Fan and Jiarui Fang and Jing Fang and Zetao Fang and Chengjian Feng and Yu Gao and Diandian Gu and Dong Guo and Hanzhong Guo and Qiushan Guo and Boyang Hao and Hongxiang Hao and Haoxun He and Jiaao He and Qian He and Tuyen Hoang and Heng Hu and Ruoqing Hu and Yuxiang Hu and Jiancheng Huang and Weilin Huang and Zhaoyang Huang and Zhongyi Huang and Jishuo Jin and Ming Jing and Ashley Kim and Shanshan Lao and Yichong Leng and Bingchuan Li and Gen Li and Haifeng Li and Huixia Li and Jiashi Li and Ming Li and Xiaojie Li and Xingxing Li and Yameng Li and Yiying Li and Yu Li and Yueyan Li and Chao Liang and Han Liang and Jianzhong Liang and Ying Liang and Wang Liao and J. H. Lien and Shanchuan Lin and Xi Lin and Feng Ling and Yue Ling and Fangfang Liu and Jiawei Liu and Jihao Liu and Jingtuo Liu and Shu Liu and Sichao Liu and Wei Liu and Xue Liu and Zuxi Liu and Ruijie Lu and Lecheng Lyu and Jingting Ma and Tianxiang Ma and Xiaonan Nie and Jingzhe Ning and Junjie Pan and Xitong Pan and Ronggui Peng and Xueqiong Qu and Yuxi Ren and Yuchen Shen and Guang Shi and Lei Shi and Yinglong Song and Fan Sun and Li Sun and Renfei Sun and Wenjing Tang and Boyang Tao and Zirui Tao and Dongliang Wang and Feng Wang and Hulin Wang and Ke Wang and Qingyi Wang and Rui Wang and Shuai Wang and Shulei Wang and Weichen Wang and Xuanda Wang and Yanhui Wang and Yue Wang and Yuping Wang and Yuxuan Wang and Zijie Wang and Ziyu Wang and Guoqiang Wei and Meng Wei and Di Wu and Guohong Wu and Hanjie Wu and Huachao Wu and Jian Wu and Jie Wu and Ruolan Wu and Shaojin Wu and Xiaohu Wu and Xinglong Wu and Yonghui Wu and Ruiqi Xia and Xin Xia and Xuefeng Xiao and Shuang Xu and Bangbang Yang and Jiaqi Yang and Runkai Yang and Tao Yang and Yihang Yang and Zhixian Yang and Ziyan Yang and Fulong Ye and Bingqian Yi and Xing Yin and Yongbin You and Linxiao Yuan and Weihong Zeng and Xuejiao Zeng and Yan Zeng and Siyu Zhai and Zhonghua Zhai and Bowen Zhang and Chenlin Zhang and Heng Zhang and Jun Zhang and Manlin Zhang and Peiyuan Zhang and Shuo Zhang and Xiaohe Zhang and Xiaoying Zhang and Xinyan Zhang and Xinyi Zhang and Yichi Zhang and Zixiang Zhang and Haiyu Zhao and Huating Zhao and Liming Zhao and Yian Zhao and Guangcong Zheng and Jianbin Zheng and Xiaozheng Zheng and Zerong Zheng and Kuan Zhu and Feilong Zuo},
      year={2026},
      eprint={2604.14148},
      archivePrefix={arXiv},
      primaryClass={cs.CV},
      url={https://arxiv.org/abs/2604.14148}, 
}

@misc{openai2024sora,
  author = {OpenAI},
  title  = {Video generation models as world simulators},
  year   = {2024},
  url    = {https://openai.com/index/video-generation-models-as-world-simulators/},
  note   = {Accessed: 2026-04-29}
}

@techreport{googleveo3report2025,
  author      = {Google DeepMind},
  title       = {Veo 3 Technical Report},
  institution = {Google},
  year        = {2025},
  url         = {https://storage.googleapis.com/deepmind-media/veo/Veo-3-Tech-Report.pdf
},
  note        = {Accessed: 2026-04-29}
}

@misc{klingteam2025klingomnitechnicalreport,
      title={Kling-Omni Technical Report}, 
      author={Kling Team and Jialu Chen and Yuanzheng Ci and Xiangyu Du and Zipeng Feng and Kun Gai and Sainan Guo and Feng Han and Jingbin He and Kang He and Xiao Hu and Xiaohua Hu and Boyuan Jiang and Fangyuan Kong and Hang Li and Jie Li and Qingyu Li and Shen Li and Xiaohan Li and Yan Li and Jiajun Liang and Borui Liao and Yiqiao Liao and Weihong Lin and Quande Liu and Xiaokun Liu and Yilun Liu and Yuliang Liu and Shun Lu and Hangyu Mao and Yunyao Mao and Haodong Ouyang and Wenyu Qin and Wanqi Shi and Xiaoyu Shi and Lianghao Su and Haozhi Sun and Peiqin Sun and Pengfei Wan and Chao Wang and Chenyu Wang and Meng Wang and Qiulin Wang and Runqi Wang and Xintao Wang and Xuebo Wang and Zekun Wang and Min Wei and Tiancheng Wen and Guohao Wu and Xiaoshi Wu and Zhenhua Wu and Da Xie and Yingtong Xiong and Yulong Xu and Sile Yang and Zikang Yang and Weicai Ye and Ziyang Yuan and Shenglong Zhang and Shuaiyu Zhang and Yuanxing Zhang and Yufan Zhang and Wenzheng Zhao and Ruiliang Zhou and Yan Zhou and Guosheng Zhu and Yongjie Zhu},
      year={2025},
      eprint={2512.16776},
      archivePrefix={arXiv},
      primaryClass={cs.CV},
      url={https://arxiv.org/abs/2512.16776}, 
}

@misc{LumaRay2,
  author = {{Luma AI}},
  title = {Ray2: Luma Ray2 Video Model},
  year = {2025},
  url = {https://lumalabs.ai/ray2},
  note = {Accessed: 2026-04-29}
}

@misc{LumaRay3,
  author = {{Luma AI}},
  title = {Ray3: Advanced Reasoning Video Model},
  year = {2025},
  url = {https://lumalabs.ai/ray},
  note = {Accessed: 2026-04-29}
}

@misc{joseph2026interpretingphysicsvideoworld,
      title={Interpreting Physics in Video World Models}, 
      author={Sonia Joseph and Quentin Garrido and Randall Balestriero and Matthew Kowal and Thomas Fel and Shahab Bakhtiari and Blake Richards and Mike Rabbat},
      year={2026},
      eprint={2602.07050},
      archivePrefix={arXiv},
      primaryClass={cs.CV},
      url={https://arxiv.org/abs/2602.07050}, 
}

@misc{jiang2024genaiarenaopenevaluation,
      title={GenAI Arena: An Open Evaluation Platform for Generative Models}, 
      author={Dongfu Jiang and Max Ku and Tianle Li and Yuansheng Ni and Shizhuo Sun and Rongqi Fan and Wenhu Chen},
      year={2024},
      eprint={2406.04485},
      archivePrefix={arXiv},
      primaryClass={cs.AI},
      url={https://arxiv.org/abs/2406.04485}, 
}

@misc{li2024genaibenchevaluatingimprovingcompositional,
      title={GenAI-Bench: Evaluating and Improving Compositional Text-to-Visual Generation}, 
      author={Baiqi Li and Zhiqiu Lin and Deepak Pathak and Jiayao Li and Yixin Fei and Kewen Wu and Tiffany Ling and Xide Xia and Pengchuan Zhang and Graham Neubig and Deva Ramanan},
      year={2024},
      eprint={2406.13743},
      archivePrefix={arXiv},
      primaryClass={cs.CV},
      url={https://arxiv.org/abs/2406.13743}, 
}

@misc{xue2025seeingarrowtimelarge,
      title={Seeing the Arrow of Time in Large Multimodal Models}, 
      author={Zihui Xue and Mi Luo and Kristen Grauman},
      year={2025},
      eprint={2506.03340},
      archivePrefix={arXiv},
      primaryClass={cs.CV},
      url={https://arxiv.org/abs/2506.03340}, 
}

@misc{hailuoai,
  author       = {MiniMax},
  title        = {Hailuo {AI} - {T}ext to {V}ideo \& {I}mage to {V}ideo Generator},
  year         = {2026},
  url          = {https://hailuoai.video/},
  note         = {Accessed: 2026-05-03},
  description  = {A platform for creating high-quality AI videos from text and images, developed by MiniMax.}
}

@misc{hacohen2024ltxvideorealtimevideolatent,
      title={LTX-Video: Realtime Video Latent Diffusion}, 
      author={Yoav HaCohen and Nisan Chiprut and Benny Brazowski and Daniel Shalem and Dudu Moshe and Eitan Richardson and Eran Levin and Guy Shiran and Nir Zabari and Ori Gordon and Poriya Panet and Sapir Weissbuch and Victor Kulikov and Yaki Bitterman and Zeev Melumian and Ofir Bibi},
      year={2024},
      eprint={2501.00103},
      archivePrefix={arXiv},
      primaryClass={cs.CV},
      url={https://arxiv.org/abs/2501.00103}, 
}

@misc{hacohen2026ltx2efficientjointaudiovisual,
      title={LTX-2: Efficient Joint Audio-Visual Foundation Model}, 
      author={Yoav HaCohen and Benny Brazowski and Nisan Chiprut and Yaki Bitterman and Andrew Kvochko and Avishai Berkowitz and Daniel Shalem and Daphna Lifschitz and Dudu Moshe and Eitan Porat and Eitan Richardson and Guy Shiran and Itay Chachy and Jonathan Chetboun and Michael Finkelson and Michael Kupchick and Nir Zabari and Nitzan Guetta and Noa Kotler and Ofir Bibi and Ori Gordon and Poriya Panet and Roi Benita and Shahar Armon and Victor Kulikov and Yaron Inger and Yonatan Shiftan and Zeev Melumian and Zeev Farbman},
      year={2026},
      eprint={2601.03233},
      archivePrefix={arXiv},
      primaryClass={cs.CV},
      url={https://arxiv.org/abs/2601.03233}, 
}

@misc{physionlabs2026galileo0,
  title        = {Galileo-0: Toward a Scalable World Critic for World Models},
  author       = {{Physion Labs}},
  year         = {2026},
  month        = apr,
  howpublished = {\url{https://physionlabs.ai/blog/galileo-0}},
  note         = {Accessed: 2026-05-04}
}

@article{yasuda2021physical,
  title={Physical event representations: Observers spontaneously impose discrete temporal structure in intuitive physical scene understanding},
  author={Yasuda, Shannon and Yates, Tyler and Yildirim, Ilker},
  journal={Journal of Vision},
  volume={21},
  number={9},
  pages={2672},
  year={2021},
  doi={10.1167/jov.21.9.2672}
}

@misc{yuan2026inferencetimephysicsalignmentvideo,
      title={Inference-time Physics Alignment of Video Generative Models with Latent World Models}, 
      author={Jianhao Yuan and Xiaofeng Zhang and Felix Friedrich and Nicolas Beltran-Velez and Melissa Hall and Reyhane Askari-Hemmat and Xiaochuang Han and Nicolas Ballas and Michal Drozdzal and Adriana Romero-Soriano},
      year={2026},
      eprint={2601.10553},
      archivePrefix={arXiv},
      primaryClass={cs.CV},
      url={https://arxiv.org/abs/2601.10553}, 
}

@misc{wan22-vae,
  title={WAN22 VAE: Video Variational Autoencoder for WAN Video Generation},
  author={WAN Model Team},
  year={2024},
  publisher={Hugging Face},
  howpublished={\url{https://huggingface.co/wan-model/wan22-vae}}
}

@misc{loshchilov2019decoupledweightdecayregularization,
      title={Decoupled Weight Decay Regularization}, 
      author={Ilya Loshchilov and Frank Hutter},
      year={2019},
      eprint={1711.05101},
      archivePrefix={arXiv},
      primaryClass={cs.LG},
      url={https://arxiv.org/abs/1711.05101}, 
}

@misc{chen2024videocrafter2overcomingdatalimitations,
      title={VideoCrafter2: Overcoming Data Limitations for High-Quality Video Diffusion Models}, 
      author={Haoxin Chen and Yong Zhang and Xiaodong Cun and Menghan Xia and Xintao Wang and Chao Weng and Ying Shan},
      year={2024},
      eprint={2401.09047},
      archivePrefix={arXiv},
      primaryClass={cs.CV},
      url={https://arxiv.org/abs/2401.09047}, 
}

@misc{kong2025hunyuanvideosystematicframeworklarge,
      title={HunyuanVideo: A Systematic Framework For Large Video Generative Models}, 
      author={Weijie Kong and Qi Tian and Zijian Zhang and Rox Min and Zuozhuo Dai and Jin Zhou and Jiangfeng Xiong and Xin Li and Bo Wu and Jianwei Zhang and Kathrina Wu and Qin Lin and Junkun Yuan and Yanxin Long and Aladdin Wang and Andong Wang and Changlin Li and Duojun Huang and Fang Yang and Hao Tan and Hongmei Wang and Jacob Song and Jiawang Bai and Jianbing Wu and Jinbao Xue and Joey Wang and Kai Wang and Mengyang Liu and Pengyu Li and Shuai Li and Weiyan Wang and Wenqing Yu and Xinchi Deng and Yang Li and Yi Chen and Yutao Cui and Yuanbo Peng and Zhentao Yu and Zhiyu He and Zhiyong Xu and Zixiang Zhou and Zunnan Xu and Yangyu Tao and Qinglin Lu and Songtao Liu and Dax Zhou and Hongfa Wang and Yong Yang and Di Wang and Yuhong Liu and Jie Jiang and Caesar Zhong},
      year={2025},
      eprint={2412.03603},
      archivePrefix={arXiv},
      primaryClass={cs.CV},
      url={https://arxiv.org/abs/2412.03603}, 
}

@misc{wang2025physcorrdualrewarddpophysicsconstrained,
      title={PhysCorr: Dual-Reward DPO for Physics-Constrained Text-to-Video Generation with Automated Preference Selection}, 
      author={Peiyao Wang and Weining Wang and Qi Li},
      year={2025},
      eprint={2511.03997},
      archivePrefix={arXiv},
      primaryClass={cs.CV},
      url={https://arxiv.org/abs/2511.03997}, 
}

@misc{chen2025physicalcoherencebenchmarkevaluating,
      title={A Physical Coherence Benchmark for Evaluating Video Generation Models via Optical Flow-guided Frame Prediction}, 
      author={Yongfan Chen and Xiuwen Zhu and Tianyu Li},
      year={2025},
      eprint={2502.05503},
      archivePrefix={arXiv},
      primaryClass={cs.CV},
      url={https://arxiv.org/abs/2502.05503}, 
}

@misc{tragoudaras2026evaluatingnewtonianmechanicsvideo,
      title={Evaluating Newtonian Mechanics in Video Generative Models with Real Physical Systems}, 
      author={Antonios Tragoudaras and Chenyu Zhang and Daniil Cherniavskii and Antonios Vozikis and Thijmen Nijdam and Derck W. E. Prinzhorn and Mark Bodracska and Nicu Sebe and Andrii Zadaianchuk and Efstratios Gavves},
      year={2026},
      eprint={2504.02918},
      archivePrefix={arXiv},
      primaryClass={cs.CV},
      url={https://arxiv.org/abs/2504.02918}, 
}

\appendix
\clearpage
\section{Appendix}
\label{app}
\subsection{More Qualitative Results}
\label{app:cases}

\begin{figure}[ht!]
    \centering
    \includegraphics[width=1.0\linewidth]{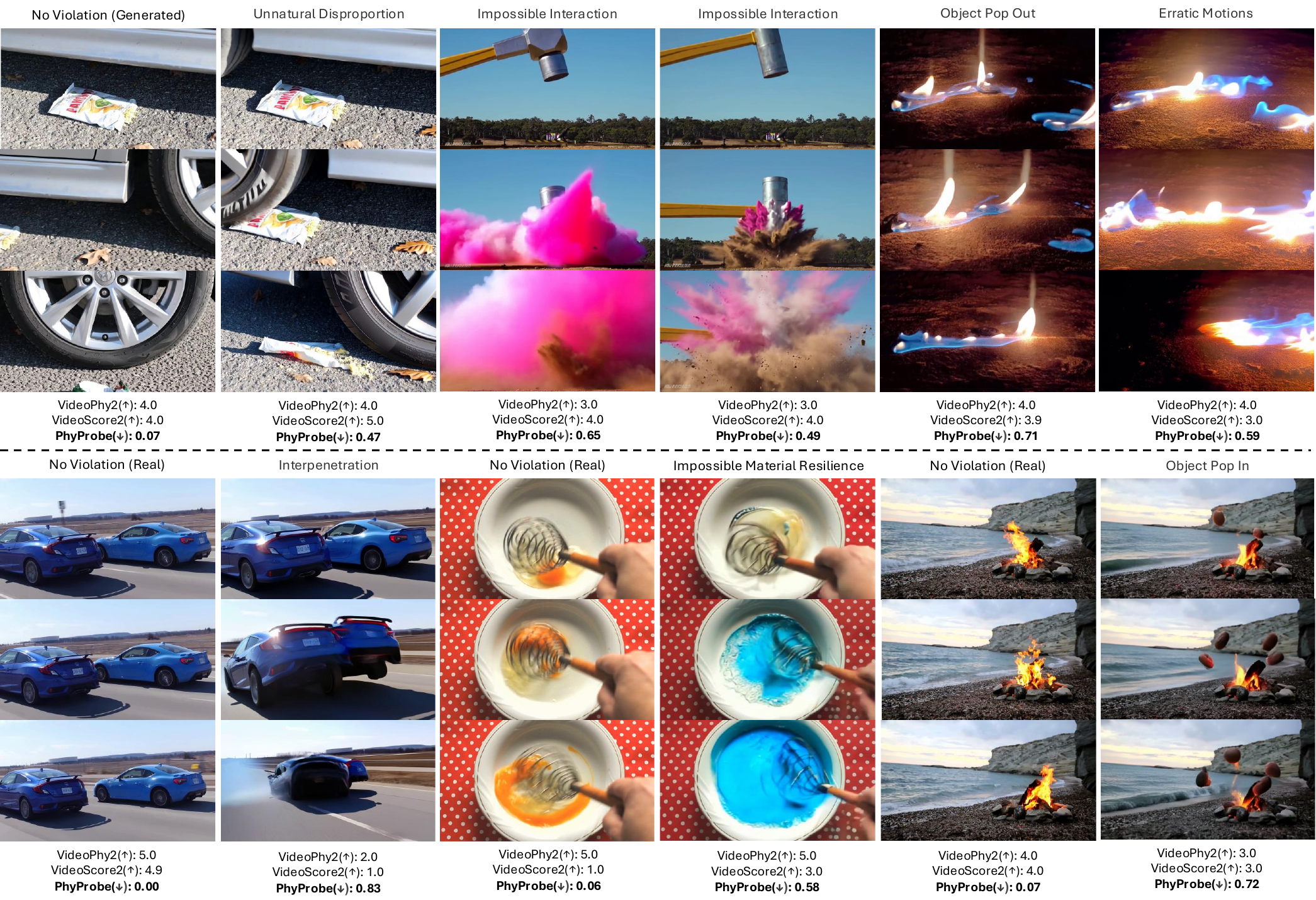}
    \caption{
    Qualitative comparison of image-to-video (I2V) generations with varying levels of physical consistency. \phyprobe{} assigns low violation scores to videos with coherent object motion and interaction dynamics, while assigning higher scores to videos exhibiting implausible behaviors such as inconsistent motion, object deformation, or physical discontinuities. These examples illustrate that \phyprobe{} captures continuous variation in physical violation severity beyond binary plausible/implausible judgments.
    }
    \label{fig:ours_qual_i2v}
\end{figure}

\begin{figure}[ht!]
    \centering
    \includegraphics[width=1.0\linewidth]{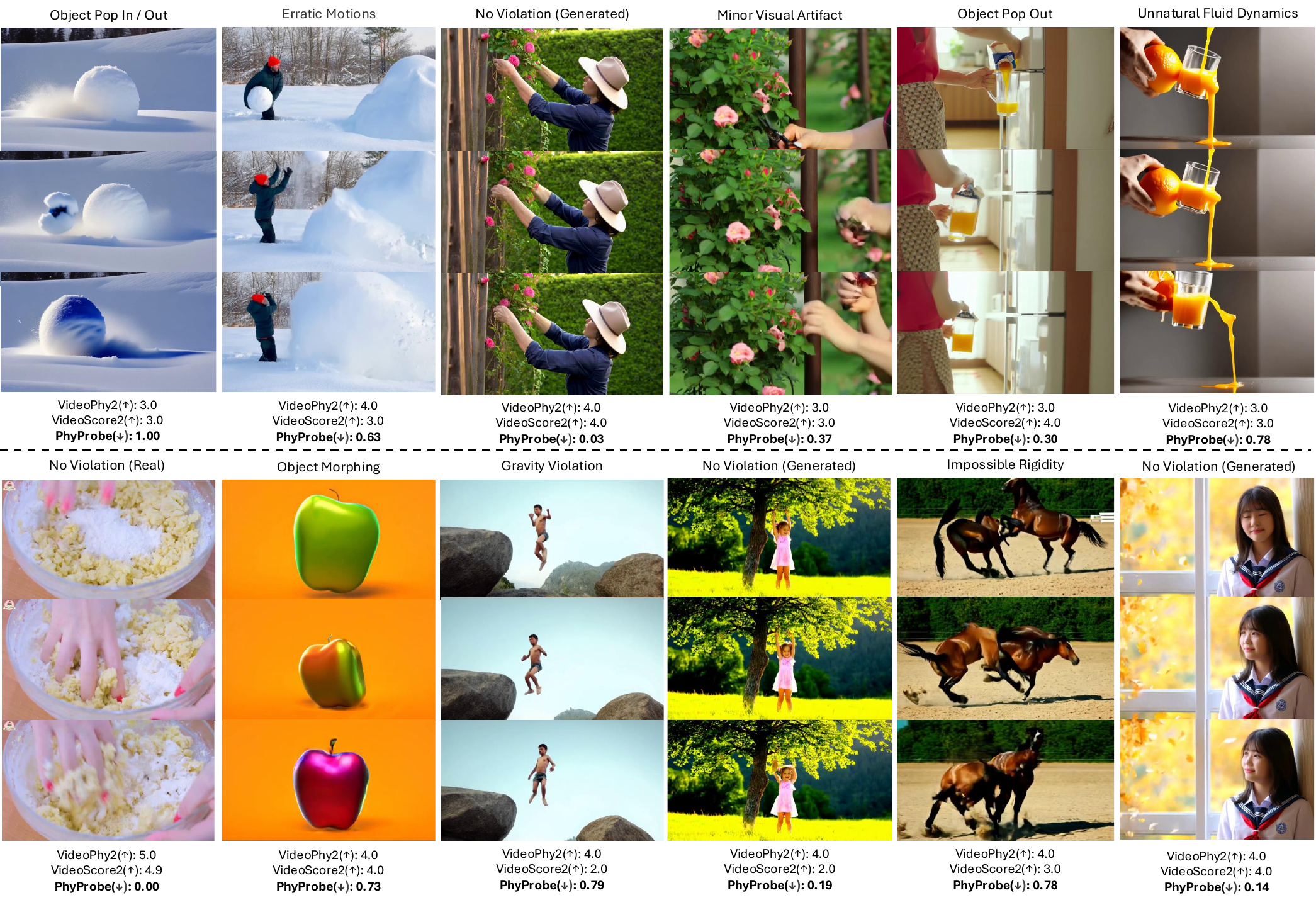}
    \caption{
    Qualitative comparison of text-to-video (T2V) generations under no semantic correspondence. Despite depicting entirely different scenes and prompts, \phyprobe{} is still able to form meaningful pairwise physical consistency comparisons by evaluating the severity of physical violations directly from video dynamics. Videos with coherent motion and interaction patterns receive lower violation scores, while videos exhibiting implausible motion, object deformation, or inconsistent dynamics receive higher scores.
    }
    \label{fig:ours_qual_t2v}
\end{figure}

Figures~\ref{fig:ours_qual_i2v} and~\ref{fig:ours_qual_t2v} present additional qualitative examples of \phyprobe{} on both image-to-video (I2V) and text-to-video (T2V) generation benchmarks. These examples illustrate that the predicted violation scores correlate well with the severity of perceived physical inconsistencies across diverse scenes, generators, and prompting settings. In particular, Figure~\ref{fig:ours_qual_t2v} demonstrates that \phyprobe{} can compare physical consistency even when videos have no semantic correspondence, supporting its use in heterogeneous pairwise evaluation.

\begin{figure}[ht!]
    \centering
    \includegraphics[width=1.0\linewidth]{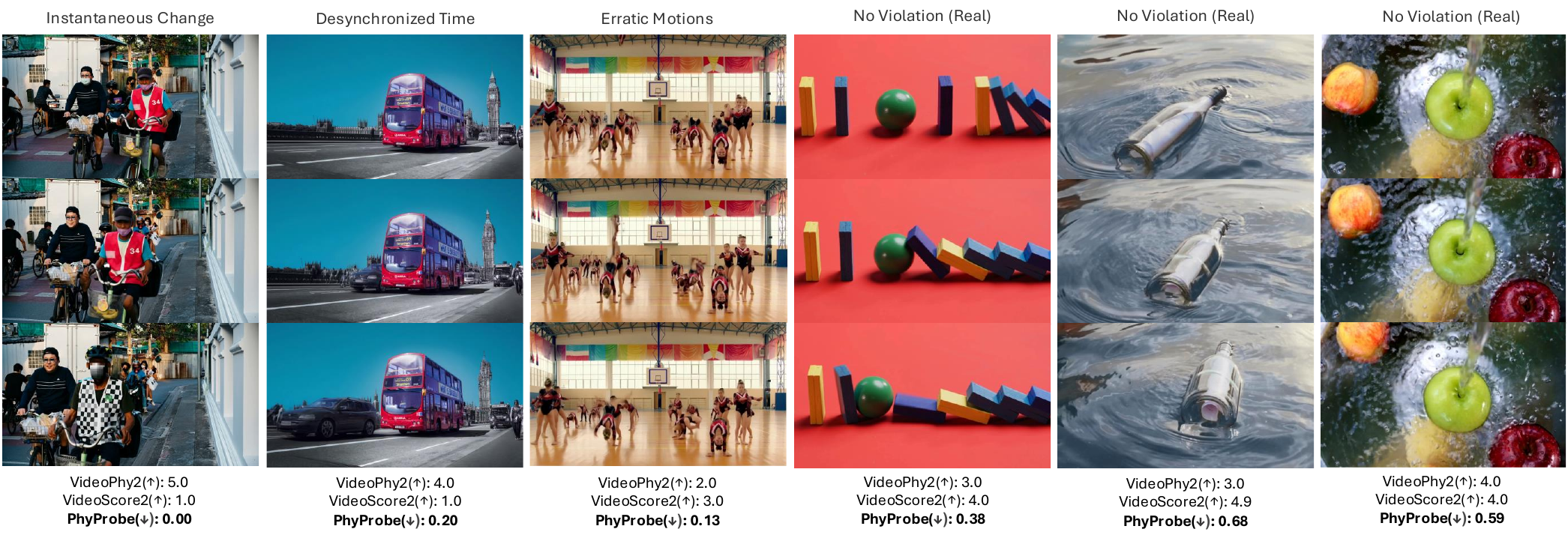}
    \caption{
    Failure cases of \phyprobe{}. In some cases, \phyprobe{} assigns elevated violation scores to real videos containing unusual or rarely observed motion patterns, suggesting limitations in the underlying representation space and training distribution. We also observe reduced sensitivity to violations that occur extremely briefly or instantaneously (e.g., sudden appearance, disappearance, or abrupt state changes), where the anomaly may not persist long enough to be strongly encoded in the aggregated video representation.
    }
    \label{fig:ours_qual_fail}
\end{figure}

Figure~\ref{fig:ours_qual_fail} presents representative failure cases. While \phyprobe{} generally captures physically implausible motion, it can occasionally assign elevated violation scores to videos containing unusual but physically valid dynamics, and may underdetect violations that occur only momentarily. These examples highlight directions for improving long-horizon temporal representations and more comprehensive physical supervision.

\clearpage
\subsection{More Ablations}
\label{app:more_ablations}

$\triangleright$ \textbf{Effect of the \phyprobe{} training objective under the same backbone.} In Table~\ref{tab:ablation_backbone_invar} we show \phyprobe{} beats V-JEPA2 Suprise baseline within the same backbone, and therefore shown \phyprobe{} does not solely attribute the gain to a stronger backbone representation.

\begin{table*}[h]
\centering
\small
\caption{We compare \phyprobe{} with V-JEPA2 suprise. Both methods use the identical V-JEPA2 ViT-H/16 backbone. \phyprobe{} consistently improves over the original V-JEPA2 Surprise baseline, indicating that the gains are not solely attributable to stronger backbone representations.}
\label{tab:ablation_backbone_invar}
\resizebox{\columnwidth}{!}{
\setlength{\tabcolsep}{4pt}
\begin{tabular}{l c cccc c cccc}
\toprule
&
&
\multicolumn{4}{c}{\textbf{Pairwise Accuracy (\%) $\uparrow$}}
&
&
\multicolumn{4}{c}{\textbf{Human Correlation $\uparrow$}} \\
\cmidrule(lr){3-6}
\cmidrule(lr){8-11}

\textbf{Method}
&
\textbf{Backbone}
&
ImplB$^p$
&
VP2$^p$
&
PhyDEx$^p$
&
VF2$^p$
&
&
$\rho_{\mathrm{VP2}}$
&
$r_{\mathrm{VP2}}$
&
$\rho_{\mathrm{VF2}}$
&
$r_{\mathrm{VF2}}$
\\
\midrule

V-JEPA2 Surprise
&
V-JEPA2 ViT-H/16
&
33.3
&
46.3
&
49.0
&
45.5
&
&
0.065
&
0.065
&
0.121
&
0.133
\\

\textbf{\phyprobe{}}
&
V-JEPA2 ViT-H/16
&
\textbf{81.4}
&
\textbf{60.2}
&
\textbf{70.7}
&
\textbf{76.3}
&
&
\textbf{0.301}
&
\textbf{0.303}
&
\textbf{0.623}
&
\textbf{0.630}
\\
\midrule
\rowcolor{gray!15}$\Delta$
&
--
&
+48.1
&
+13.9
&
+21.7
&
+30.8
&
&
+0.236
&
+0.238
&
+0.502
&
+0.497
\\
\bottomrule
\end{tabular}
}
\end{table*}

$\triangleright$ \textbf{Effect of the minimum score gap in evaluation set design.} We further analyze the effect of the minimum ground-truth score gap used to construct the evaluation pairs. As shown in Table~\ref{tab:videophy2_gap_threshold}, \phyprobe{}\ consistently outperforms existing baselines across different minimum-gap thresholds, demonstrating that its advantage is robust to variations in evaluation set construction rather than being tied to a particular pair-selection criterion.

\begin{table}[h]
\centering
\caption{
Pairwise accuracy (\%) on the \textbf{text-correspondence subset} of
VP2$^p$ under different minimum ground-truth score-gap thresholds.
Only video pairs generated from the same text prompt are considered,
following the pairwise evaluation protocol; equal-score pairs are excluded.
The same checkpoint for each method is evaluated across all rows, while only
the evaluation subset is changed.
}
\label{tab:videophy2_gap_threshold}
\resizebox{\columnwidth}{!}{
\begin{tabular}{c|c|cccc}
\toprule
\textbf{Minimum} &
\textbf{\# Test} &
\multicolumn{4}{c}{\textbf{Pairwise Accuracy (\%)}} \\
\cmidrule(lr){3-6}
\textbf{Score Gap} &
\textbf{Pairs} &
\textbf{VideoScore} &
\textbf{VideoPhy2-AutoEval} &
\textbf{V-JEPA2 Surprise} &
\textbf{\phyprobe{}} \\
\midrule
$\Delta_{\mathrm{eval}} \geq 1$ & 1280 & 49.8 & 28.5 & 46.3 & \textbf{66.1} \\
$\Delta_{\mathrm{eval}} \geq 2$ & ~~628 & 52.0 & 32.8 & 44.6 & \textbf{71.2} \\
$\Delta_{\mathrm{eval}} \geq 3$ & ~~202 & 52.0 & 34.2 & 46.0 & \textbf{77.2} \\
$\Delta_{\mathrm{eval}} \geq 4$ & ~~~35 & 37.1 & 45.7 & 48.6 & \textbf{74.3} \\
\bottomrule
\end{tabular}
}
\end{table}

$\triangleright$ \textbf{Leave-one-dataset-out (LODO) analysis.}
To evaluate whether \phyprobe{}\ relies on a particular training source, we retrain the model while excluding TRAVL, VideoPhy2, or VideoFeedback2 from training. As shown in Table~\ref{tab:lodo}, performance degrades on the held-out benchmark and on VP2$^p$, while remaining strong on the others. The results show that useful signals transfer, but the substantial corresponding-benchmark drops also demonstrate meaningful dataset-specific dependence.

\begin{table}[h]
\centering
\caption{\textbf{Leave-one-dataset-out (LODO) analysis.}
We retrain \phyprobe{}\ while excluding one training dataset (TRAVL, VideoPhy2, or VideoFeedback2) at a time and evaluate pairwise accuracy (\%) across four physical video benchmarks. While performance degrades on the corresponding held-out benchmark and, in every case, on VP2$^p$; it remains strong on the others, indicating partial transfer across sources alongside meaningful dependence on each dataset.}
\label{tab:lodo}

\begin{tabular}{lcccc}
\toprule
\textbf{Method}
& \textbf{ImplB$^p$}
& \textbf{VP2$^p$}
& \textbf{PhyDEx$^p$}
& \textbf{VF2$^p$} \\
\midrule

VP2-AutoEval
& 28.0
& 28.5
& 30.5
& 36.3 \\

V-JEPA2 Surprise
& 33.3
& 46.3
& 49.0
& 45.5 \\

VideoScore2
& 68.7
& 49.7
& 60.6
& 65.0 \\

\midrule

\textbf{\phyprobe{}}
& \textbf{98.0}
& \textbf{66.1}
& \textbf{98.9}
& \textbf{75.2} \\

\midrule

\textbf{\phyprobe{} (w/o TRAVL)}
& 92.1
& 58.7
& 98.2
& 74.7 \\

\rowcolor{gray!15}$\Delta$ (w/o TRAVL $-$ full)
& $-5.9$
& $-7.4$
& $-0.7$
& $-0.5$ \\
		
\textbf{\phyprobe{} (w/o VideoPhy2)}
& 96.1
& 58.3
& 99.3
& 69.5 \\
		
\rowcolor{gray!15}$\Delta$ (w/o VideoPhy2 $-$ full)
& $-1.9$
& $-7.8$
& $+0.4$
& $-5.7$ \\

\textbf{\phyprobe{} (w/o VideoFeedback2)}
& 92.8
& 57.1
& 99.7
& 58.0 \\
		
\rowcolor{gray!15}$\Delta$ (w/o VideoFeedback2 $-$ full)
& $-5.2$
& $-9.0$
& $+0.8$
& $-17.2$ \\

\bottomrule
\end{tabular}
\end{table}

$\triangleright$ \textbf{Effect of score normalization.}
We compare the proposed violation-flooring normalization with a simple linear normalization while keeping all other training settings unchanged. As shown in Table~\ref{tab:linear_mapping}, the proposed normalization consistently improves pairwise accuracy across all four physical video benchmarks, suggesting that preserving greater score resolution in the low-violation regime provides more effective supervision. As preliminary supporting evidence, we also conduct a small blinded human preference study comparing the predicted score differences produced by the two normalization schemes. As shown in Table~\ref{tab:human_pref}, annotators leaned toward the proposed normalization.

\begin{table}[h]
\centering
\caption{\textbf{Effect of score normalization on PHYPROBE.}
We compare the proposed violation-flooring normalization against a simple linear normalization while keeping all other training settings identical. Results report pairwise accuracy (\%) on four physical video benchmarks.}
\label{tab:linear_mapping}

\begin{tabular}{lcccc}
\toprule
\textbf{Method}
& \textbf{ImplB$^p$}
& \textbf{VP2$^p$}
& \textbf{PhyDEx$^p$}
& \textbf{VF2$^p$} \\
\midrule

\textbf{\phyprobe (linear)}
& 96.7
& 60.0
& 96.5
& 71.0 \\

\textbf{\phyprobe}
& \textbf{98.0}
& \textbf{66.1}
& \textbf{98.9}
& \textbf{75.2} \\

\rowcolor{gray!15}$\Delta$ (linear $-$ proposed)
& $-1.3$
& $-6.1$
& $-2.4$
& $-4.2$ \\

\bottomrule
\end{tabular}
\end{table}

\begin{table}[h]
\centering
\caption{\textbf{Blinded human preference evaluation of scalar target normalization.}
We compare the proposed violation-flooring normalization against a simple linear normalization using a blinded preference evaluation. Annotators were shown the predicted score differences from both models in random order and selected which better reflected the perceived difference in physical violation severity. Each of the $n$ pairs was independently rated by four non-expert annotators; percentages are pooled over all annotator judgments. Given the small sample ($n=30$ pairs) and inter-rater agreement of Fleiss' $\kappa = 0.148$, we regard this as preliminary supporting evidence of a preference trend.}
\label{tab:human_pref}

\begin{tabular}{lcc}
\toprule
\textbf{Subset}
& \textbf{Linear}
& \textbf{Proposed} \\
\midrule

ImplB$^p$ ($n=10$)
& 17.5\%
& \textbf{82.5\%} \\

VF2$^p$ ($n=10$)
& 37.5\%
& \textbf{62.5\%} \\

VP2$^p$ ($n=10$)
& 25.0\%
& \textbf{75.0\%} \\

\midrule

Overall ($n=30$)
& 26.7\%
& \textbf{73.3\%} \\

\midrule

Fleiss' $\kappa$
& \multicolumn{2}{c}{0.148 (4 raters)} \\

Unanimous agreement
& \multicolumn{2}{c}{14/30 pairs (46.7\%)} \\

\bottomrule
\end{tabular}
\end{table}

\begin{figure}[h]
    \centering

    \begin{subfigure}[t]{0.245\linewidth}
        \centering
        \includegraphics[width=\linewidth]{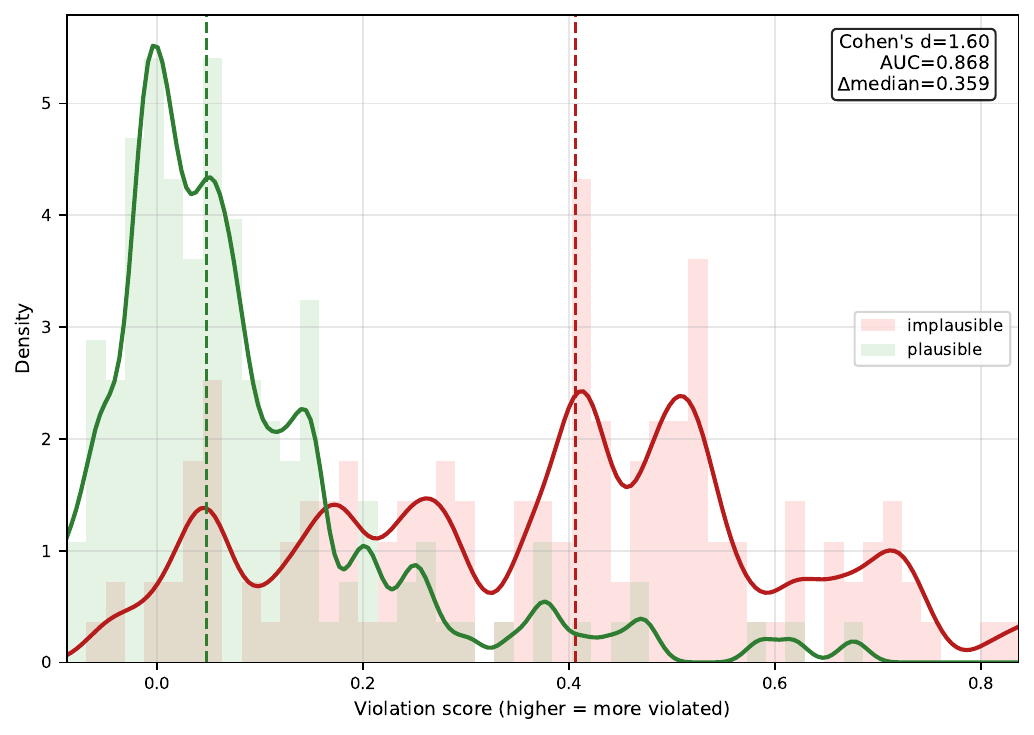}
        \caption{\phyprobe on ImplB$^p$}
    \end{subfigure}
    \hfill
    \begin{subfigure}[t]{0.245\linewidth}
        \centering
        \includegraphics[width=\linewidth]{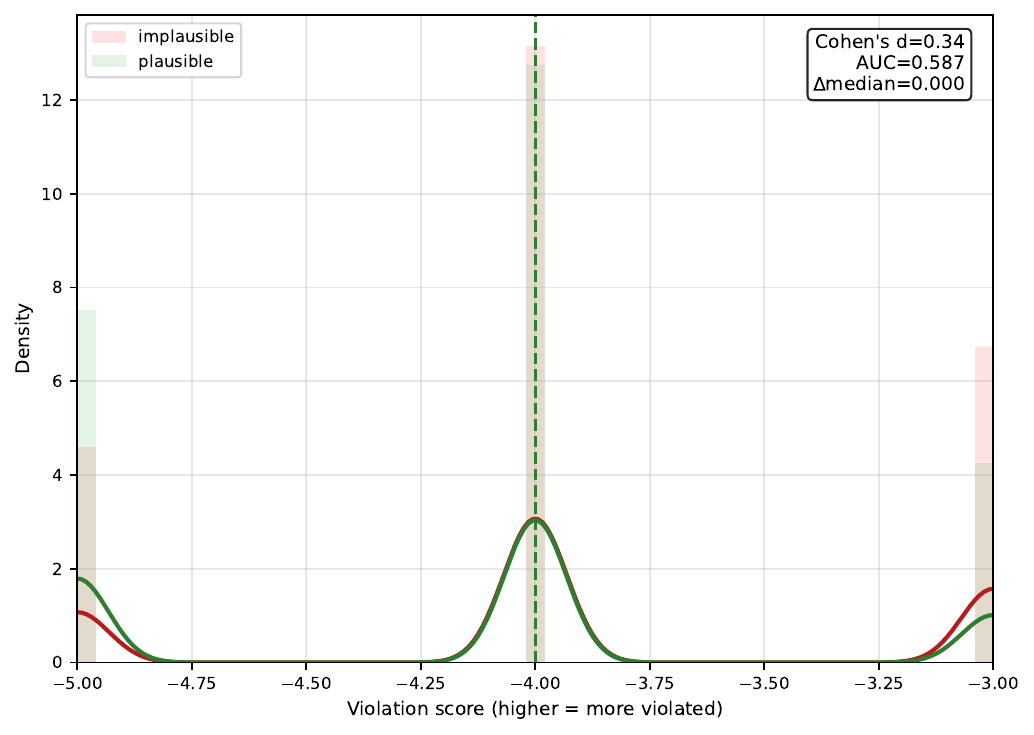}
        \caption{VP2 on ImplB$^p$}
    \end{subfigure}
    \hfill
    \begin{subfigure}[t]{0.245\linewidth}
        \centering
        \includegraphics[width=\linewidth]{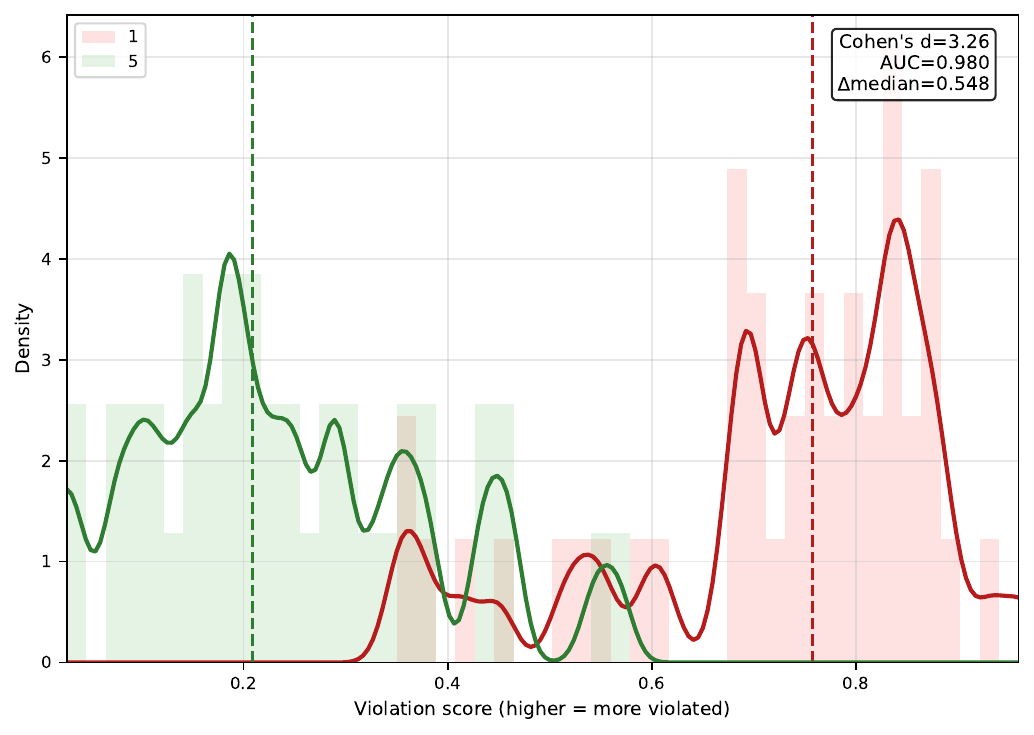}
        \caption{\phyprobe on VF2$^p$}
    \end{subfigure}
    \hfill
    \begin{subfigure}[t]{0.245\linewidth}
        \centering
        \includegraphics[width=\linewidth]{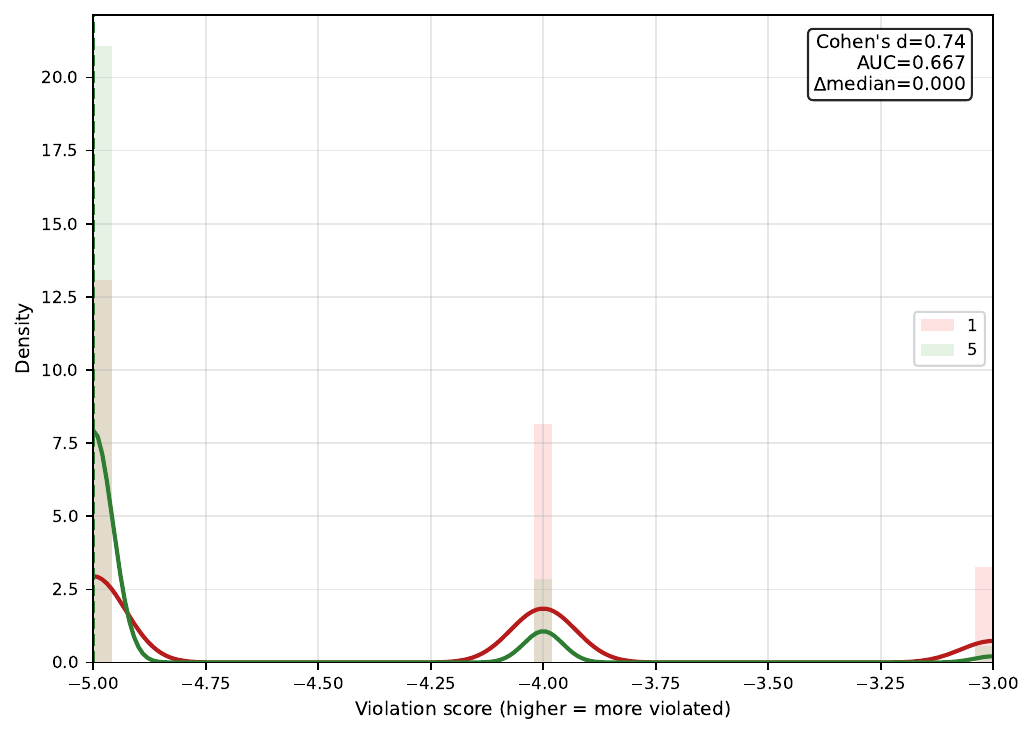}
        \caption{VP2 on VF2$^p$}
    \end{subfigure}

    \caption{\textbf{Predicted score distributions for plausible and implausible videos.}
    Score distributions produced by \phyprobe\ and VideoPhy2-AutoEval (VP2) on ImplB$^p$ (a,b) and VF2$^p$ (c,d). For VF2$^p$, only samples with human scores of 1 and 5 are shown, corresponding to clearly implausible and plausible videos, respectively, to reduce ambiguity from intermediate ratings.}
    \label{fig:dist-vis}
\end{figure}

$\triangleright$ \textbf{Distribution of predicted scores.}
Figure~\ref{fig:dist-vis} compares the predicted score distributions of \phyprobe{}\ and VideoPhy2-AutoEval on ImplB$^p$ and VF2$^p$. \phyprobe{}\ produces substantially better separation between plausible and implausible videos, whereas the baseline exhibits considerable overlap between the two distributions. This indicates that the learned violation scores provide improved discriminability for physical plausibility.

\begin{table}[t]
\centering
\small
\caption{Sensitivity of \phyprobe{} ``w/ tie'' accuracy (\%) to the model-side tie threshold, expressed as a percentile of absolute score differences on each benchmark. The 20th percentile is used for \phyprobe{} results in Table~\ref{tab:quality_pairwise}. ``w/o tie'' columns compare raw scores without a threshold and are unaffected.}
\label{tab:tie_sensitivity}
\begin{tabular}{lccccc}
\toprule
\textbf{Percentile} & 5 & 10 & 20 & 30 & 40 \\
\midrule
MonetBench          & 69.8 & 71.4 & \textbf{72.7} & 71.8 & 72.4 \\
RewardBench Overall & 63.2 & 63.6 & 64.1 & 64.6 & \textbf{64.8} \\
RewardBench MQ      & 57.9 & 59.4 & 62.3 & 65.0 & \textbf{67.9} \\
RewardBench VQ      & 58.0 & 59.3 & 61.4 & 63.4 & \textbf{65.6} \\
\bottomrule
\end{tabular}
\end{table}

\subsection{Definition of Physical Consistency}

\textbf{"Physical Consistency" can be defined from different perspectives}. Here we follow the operational definition adopted by existing public benchmarks~\citep{bansal2024videophyevaluatingphysicalcommonsense, he2025videoscore2thinkscoregenerative, riochet2020intphysframeworkbenchmarkvisual}, where it is approximated by perceived physical plausibility. While the exact operationalization varies slightly across datasets, they all rely on human judgments of whether physical laws appear to be violated. This variability further motivates aggregating heterogeneous supervision from multiple complementary sources under a unified training objective. Since physical plausibility and visual quality are often intertwined, human judgments naturally reflect both aspects. In essence, PhyProbe is trained to approximate the degree to which a human would perceive physical laws to be violated.

\textbf{Common Video Abnormalities.} We categorize common abnormalities in generated videos into four groups. Although these categories frequently co-occur with perceived physical inconsistency, not every instance necessarily constitutes a violation of physical laws.

\begin{itemize}
    \item \textbf{Perception abnormalities} affect visual fidelity and appearance without necessarily violating physical laws, including artifacts such as noise, flickering lighting, repeated scenery, missing details, and unnatural color palettes.  
    \item \textbf{Objectness abnormalities} relate to the existence, shape, and spatial integrity of entities, such as objects appearing or disappearing, distorted proportions, impossible poses, sudden morphing, or inconsistent spatial layouts. 
    \item \textbf{Behavioral abnormalities} capture physically implausible interactions and material properties, such as object interpenetration, unrealistic fluid dynamics, impossible deformation or resilience, and erratic motion that violates expected force and collision dynamics. 
    \item \textbf{Time abnormalities} involve disruptions in temporal consistency, including desynchronized motion, instantaneous corrections, abrupt start or stop behaviors, and violations of causal ordering. 
\end{itemize}

\clearpage

\section{Implementation Details}
\label{app:implementation_detail}

\subsection{Dataset Construction}
\label{app:b_dataset}

\textbf{Dataset Sourcing.}
We constructed a diverse dataset from prior video physics evaluation works, including VideoScore2~\citep{he2025videoscore2thinkscoregenerative}, VideoPhy2~\citep{bansal2025videophy2challengingactioncentricphysical}, BrokenVideos~\citep{lin2025brokenvideosbenchmarkdatasetfinegrained}, PAI-Bench~\citep{zhou2025paibenchcomprehensivebenchmarkphysical}, TRAVL~\citep{motamed2025travlrecipemakingvideolanguage}, and ImpossibleVideos~\citep{bai2025impossiblevideos}. These datasets contains generated videos originate from a diverse set of modern video generation systems, including proprietary and open-source Text-to-Video (T2V) and Image-to-Video (I2V) models, such as CogVideoX~\citep{yang2025cogvideoxtexttovideodiffusionmodels}, Wan2.1~\citep{wan2025wanopenadvancedlargescale}, Sora~\citep{openai2024sora}, Ray2~\citep{LumaRay2}, VideoCrafter2~\citep{chen2024videocrafter2overcomingdatalimitations}, Veo~\citep{googleveo3report2025}, etc.~\citep{klingteam2025klingomnitechnicalreport, nvidia2025cosmosworldfoundationmodel, hailuoai, kong2025hunyuanvideosystematicframeworklarge}. Such diversity reduces the reliance on model-specific artifacts.

\textbf{Data Preprocessing.} Each video is resized to a fixed square resolution while preserving the original aspect ratio via letterboxing, filling the empty bars with mid-gray (pixel value 0.5). Frames are sampled uniformly in time at 16 fps, yielding $T=48$ frames per clip.

\begin{figure}[!h]
    \centering
    \includegraphics[width=0.9\linewidth]{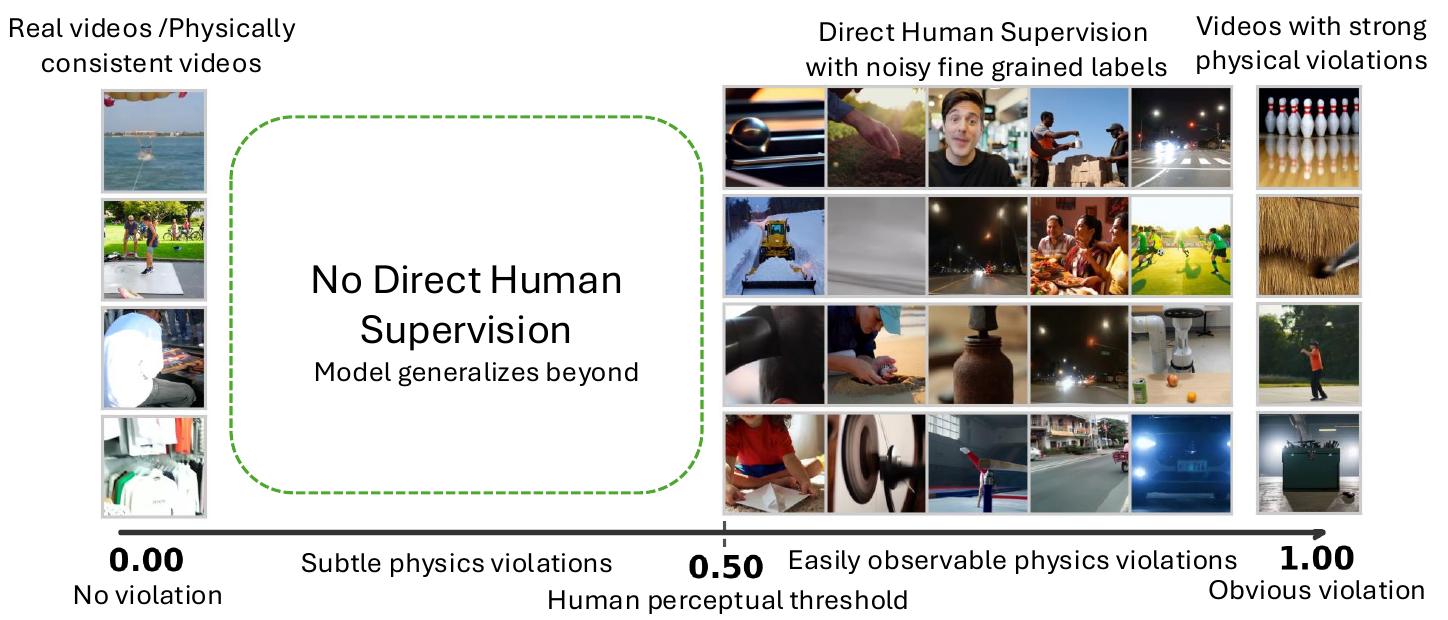}
    \caption{
    Illustration of the supervision regime for physical plausibility evaluation. Human annotations primarily capture clearly observable violations (scores $\geq 0.5$), while the region below the perceptual threshold remains largely unsupervised. Importantly, scores in the range $[0,0.5)$ do not indicate the absence of violations, but rather subtle or weakly perceptible physical inconsistencies. \phyprobe{} produces differentiated scores in this region, but their interpretation remains unvalidated due to the absence of direct severity annotations. Figure~\ref{fig:corr-vis} and Table~\ref{tab:linear_mapping} provide indirect evidence that finer resolution in the low-violation regime improves supervision.
    }
    \label{fig:training_regime}
\end{figure}

\textbf{Score Assignment and Normalization.}
For VideoScore2~\citep{he2025videoscore2thinkscoregenerative} and VideoPhy2~\citep{bansal2025videophy2challengingactioncentricphysical}, we assume the 1--5 Likert scales for physical plausibility share a comparable semantic interpretation across datasets. Since \phyprobe{} models violation severity rather than plausibility, we invert and normalize these annotations into the range $[0,1]$, where $0$ denotes physically consistent motion and $1$ denotes severe physical violation. Motivated by the thresholded nature of human perception reported in related settings~\citep{nightingale2017identify,nightingale2019detect,nightingale2022ai} , we apply a violation-flooring scheme; the specific mapping is a design choice, supported empirically by the ablation in Table~\ref{tab:linear_mapping} and, preliminarily, by the small preference study in Table~\ref{tab:human_pref}. Specifically, PC=5 is mapped to $0.0$, PC=4 to $0.5$, PC=3 to $0.667$, PC=2 to $0.833$, and PC=1 to $1.0$. Under this formulation, videos judged as largely plausible by humans may still contain subtle violations below the perceptual threshold, while increasingly implausible ratings are compressed toward the high-violation regime. For BrokenVideos~\citep{lin2025brokenvideosbenchmarkdatasetfinegrained}, which provides spatial-temporal mask annotations for visual and physical errors, we derive approximate violation scores from mask statistics including spatial area, temporal persistence, and proximity to the image center. Since these annotations primarily correspond to visibly salient failures, we apply a similar flooring strategy and map the resulting scores to the range $[0.5, 0.9]$. For Kinetics-700~\citep{carreira2022shortnotekinetics700human}, PAI-Bench~\citep{zhou2025paibenchcomprehensivebenchmarkphysical}, TRAVL~\citep{motamed2025travlrecipemakingvideolanguage}, and ImpossibleVideos~\citep{bai2025impossiblevideos}, we use anchor-based supervision. Real videos are assigned a target score of $0$, while synthetic failure cases from ImpossibleVideos and TRAVL are anchored toward $1$, providing approximate endpoints for calibration of the violation scale. The overall supervision regime is illustrated in Figure~\ref{fig:training_regime}.

\subsubsection{Construction Detail of Training Datasets and Supervision Labels}
\label{sec:app_train_dataset}

The main paper summarizes which supervision signals are provided by each dataset (Table \ref{tab:training_data}). Here, we describe how the corresponding supervision labels are constructed and how each dataset is converted into training examples.

\begin{itemize}

\item \textbf{PAI-Bench.}
PAI-Bench~\citep{zhou2025paibenchcomprehensivebenchmarkphysical} contains Image-to-Video model-generated videos paired with human-recorded reference videos \textbf{sharing the same image prompt input}. For each reference video, we sample up to 50 model-generated videos as negative partners, re-sampling across training epochs to diversify pair coverage. The reference video is always assigned as the winner, providing a preference learning (\textbf{rank signal}) that pushes model-generated videos below their human counterparts. We also anchor the reference videos with \textbf{calibration signal} as reference videos are physically consistent.

\item \textbf{VideoPhy2.}
VideoPhy2~\citep{bansal2025videophy2challengingactioncentricphysical} annotates videos with a physical correctness (PC) score on a 1--5 integer scale.
We form pairs \textbf{within the same action category} and assign the video with the higher PC score as the winner, treating the raw scores as continuous regression targets (\textbf{magnitude signal}) and also preference learning targets (\textbf{rank signal}).

\item \textbf{VideoFeedback2.}
VideoFeedback2~\citep{he2025videoscore2thinkscoregenerative} similarly provides physical correctness (PC) score (1--5) at the prompt level. We pair videos \textbf{without image or text prompt correspondence} and retain only pairs whose PC scores differ by at least 1, filtering out near-tie comparisons whose ordering is less reliable. The video with the higher PC score is the winner, treating the raw scores as continuous regression targets (\textbf{magnitude signal}) and also preference learning targets (\textbf{rank signal}).

\item \textbf{BrokenVideos.}
BrokenVideos~\citep{lin2025brokenvideosbenchmarkdatasetfinegrained} feature videos with collective failures, ranging from visuals to physical violations. It also contains videos exhibiting visual artifacts detected via segmentation masks. We derive a composite anomaly score $s = 0.6 \cdot \text{mask\_area} + 0.3 \cdot \text{center\_proximity} + 0.1 \cdot \text{mask\_presence\_ratio}$, where lower scores indicate fewer artifacts. We create pairs \textbf{without image or text prompt correspondence} and retain a score gap $\geq 0.4$, which exposes the model to a broader distribution of artifact severity, treating the scores as continuous regression targets (\textbf{magnitude signal}) and also preference learning targets (\textbf{rank signal}).


\item \textbf{ImpossibleVideos.}
ImpossibleVideos~\citep{bai2025impossiblevideos} features counterfactual scenes that violate physical, biological, geographical laws, and it contains physically implausible AI-generated videos against real-world videos. Pairs are then formed globally between plausible and implausible videos \textbf{without image or text prompt correspondence}, with the plausible video always assigned as the winner. It act as a preference learning pair to provide \textbf{rank signal} and we also anchor the plausible videos and implausible videos with \textbf{calibration signal} as plausible videos are physically consistent and implausible videos are systematic failures.

\item \textbf{TRAVL.}
TRAVL~\citep{motamed2025travlrecipemakingvideolanguage} contains videos annotated via multiple question-answer pairs, each carrying a per-question plausibility label (0 = plausible, 1 = implausible). We derive a per-video plausibility label by majority vote across all QA pairs, discarding videos with tied votes. Pairs are then formed globally between plausible and implausible videos \textbf{without image or text prompt correspondence}, with the plausible video always assigned as the winner. It act as a preference learning pair to provide \textbf{rank signal} and we also anchor the plausible videos and implausible videos with \textbf{calibration signal} as plausible videos are physically consistent and implausible videos are systematic failures. This dataset is complementary to ImpossibleVideos: both provide a binary plausible-vs-implausible signal, but TRAVL's labels are grounded in explicit physical reasoning questions rather than direct human annotation.

\item \textbf{Kinetics-700.}
We form pairs \textbf{within the same action category} in the Kinetics-700 dataset~\citep{carreira2022shortnotekinetics700human}. Since both videos are real and physically consistent, pairs are treated as ties under the \textbf{ranking signal}. This provides a real-video anchor for the model to learn the \textbf{calibration signal}.

\end{itemize}

\subsubsection{Construction Detail of Evaluation Datasets and Ground Truth Labels}
\label{sec:app_eval_dataset}

For every benchmark below, the model scores each video independently; pairwise accuracy and correlation are computed from those per-video scores. For benchmarks with human tie labels, \phyprobe{} declares a tie when $|s(v_A) - s(v_B)|$ falls below the 20th percentile of absolute score differences on that benchmark. The same percentile is used for all benchmarks; its effect is reported in Table~\ref{tab:tie_sensitivity}. The ``w/o tie'' columns of Table~\ref{tab:quality_pairwise} compare raw scores without a threshold and are unaffected by it.

\paragraph{Physical Domain}

\begin{itemize}

\item \textbf{ImplB$^p$.} The official ImplausiBench~\citep{motamed2025travlrecipemakingvideolanguage} contains 300 videos (150 real and 150 generated by image-to-video models conditioned on the real video's first frame) with gold-standard human judgments. We pair the real and its corresponding generated one to form correspondence pairs. 

\item \textbf{VP2$^p$.} We derive the pairwise dataset from the official VideoPhy2-test dataset~\citep{bansal2025videophy2challengingactioncentricphysical}. Each video in official VideoPhy2-test has a rating between 1 to 5 rated by a human annotator in physical commonsense. We kept text prompt corresponding pairs based on rating gap $\geq$ 1 as pairs. Equal-score pairs are excluded. Total 1280 pairs. 

\item \textbf{PhyDEx$^p$.} We derive the pairwise dataset from PhyDetEx's PID-test dataset~\citep{wang2025phydetexdetectingexplainingphysical} which contains 250 real and 250 generated videos. We sample and form 2000 pairs based on a real video paired with a generated one as pairs without image or text correspondence. 

\item \textbf{VF2$^p$.} We derive the pairwise dataset from the official VideoFeedback2-test dataset~\citep{he2025videoscore2thinkscoregenerative}. Each video has a rating between 1 to 5 rated by a human annotator in physical commonsense. We form pairs based on rating gap $\geq$ 1 as pairs without image or text correspondence. Equal-score pairs are excluded. 2000 pairs are formed. 

\item \textbf{Human Correlation VP2.} The official VideoPhy2-test set released by VideoPhy2 is used for human correlation~\citep{bansal2025videophy2challengingactioncentricphysical}. Total instances 3397, each video has a rating between 1 to 5 rated by a human annotator in physical commonsense. We use model to score video individually for computing the human correlation.

\item \textbf{Human Correlation VF2.} The official VideoFeedback2-test released by VideoScore2 is used for human correlation~\citep{he2025videoscore2thinkscoregenerative}. Total instances 500, each video has a rating between 1 to 5 rated by a human annotator in physical commonsense. We use model to score video individually for computing the human correlation.

\end{itemize}

\paragraph{Quality Domain}

\begin{itemize}

\item \textbf{MonetBench.} MonetBench is the official pairwise test set from VisionReward~\citep{xu2026visionrewardfinegrainedmultidimensionalhuman}, which each pair is a human preference pair. 

\item \textbf{Rapidata-I2V.} Rapidata-I2V is a pairwise preference dataset released on HuggingFace datasets by Rapidata~\citep{rapidata_hailuo02_marey_i2v_2025, rapidata_seedance1pro_i2v_2025}, which each pair is a human preference pair. The videos come from recent generators such as Seedance1Pro and Hailuo 2.0. 

\item \textbf{RewardBench.} VideoGen-RewardBench is the official pairwise test set from VideoReward~\citep{liu2025improvingvideogenerationhuman}, which each pair is a human preference pair. 

\end{itemize}

\subsection{Model Training}
\label{app:b_training}

\textbf{Training.} We use pretrained video encoders as feature extractor, candidates include Wan 2.2 VAE~\citep{wan22-vae}, VJEPA2~\citep{assran2025vjepa2selfsupervisedvideo}, and Perception Encoder~\citep{bolya2025perceptionencoderbestvisual}. We found Perception Encoder Core G14-448 gives the best results. The encoder is kept frozen during training. On top of the representation, we train a lightweight scoring head $g(\cdot)$ implemented as linear layers with statistic aggregation head. The scoring head outputs an unconstrained value $\tilde{s}(v) \in \mathbb{R}$; the reported violation score is $s(v) = \text{clip}(\tilde{s}(v), 0, 1)$, applied at inference only. Training is performed using the combined objective Equation~\ref{eq:loss_total}. We set $\lambda_{\text{mag}} = 10$ and $\lambda_{\text{cal}} = 100$. \phyprobe{} can be trained within 24 hours using 4 NVIDIA H100 GPUs using batch size = 8 for 10K steps. To balance heterogeneous data sources, we apply dataset-level sampling weights to prevent large datasets from dominating the learning signal. In detail, per-step sampling probabilities are: PAI-Bench 0.059, VideoPhy2 0.176, VideoFeedback2 0.294, BrokenVideos 0.118, ImpossibleVideos 0.059, TRAVL 0.176, Kinetics-700 0.118.

\textbf{Training Techniques.} To stabilize training, we adopt several complementary strategies. The projection head is optimized with AdamW~\citep{loshchilov2019decoupledweightdecayregularization} with learning $3\times10^{-4}$, weight decay = 0.01, and with a linear warmup over 500 steps followed by cosine annealing to zero over 10K total steps. To obtain a stable inference model, we maintain an Exponential Moving Average (EMA) of the projection head weights with decay 0.9999, and use the EMA checkpoint for all evaluations. When computing the ranking loss, we apply label smoothing whose strength depends on the annotated score gap. For a pair $(v_i, v_j)$ with gap $g_{ij} = |y_{v_i} - y_{v_j}|$, we set $c_{ij} = \max(0.1,\, g_{ij} / \max_{(k,l)} g_{kl})$, with the maximum over pairs from the same dataset in the current batch, and use smoothing $\epsilon_{ij} = 0.10 - 0.05\,c_{ij} \in [0.05, 0.095]$: the largest-gap pair in a batch slice receives $\epsilon=0.05$, and the floor on $c_{ij}$ keeps every pair contributing, with near-tied pairs smoothed at $\epsilon=0.095$. Pairs without scalar annotations use $c_{ij}=1$. Regression targets use a Huber loss with $\delta = 0.5$ to reduce sensitivity to annotation noise, while the hard anchor targets use MSE with a higher loss coefficient to enforce the score range boundaries firmly. Finally, the pair pool for each dataset is resampled every 2.5K steps to expose the model to a fresh set of contrasting video pairs throughout training.

\subsection{Evaluation}
\label{app:b_evaluation}

\textbf{Baselines.} All baselines, including VideoPhy2~\citep{bansal2025videophy2challengingactioncentricphysical}, VideoScore2~\citep{he2025videoscore2thinkscoregenerative}, VisionReward~\citep{xu2026visionrewardfinegrainedmultidimensionalhuman}, VideoReward~\citep{liu2025improvingvideogenerationhuman}, were directly sourced from their publicly released code on github and their provided prompt template. For VJepa2-Suprise score, we follow the implementation as described in WMReward~\citep{yuan2026inferencetimephysicsalignmentvideo} using the pretained VJEPA-2 model~\citep{assran2025vjepa2selfsupervisedvideo}. For Gemini-3.1-Flash-Lite, we use the offical API call and adopt the prompt template from Figure~\ref{fig:physical-consistency-pairwise-prompt} for accuracy and Figure~\ref{fig:physical-consistency-score-prompt} for correlation.

As some of the baselines contain scores in multiple aspects, we pick the scores that is related with physical consistency or visual quality. We compared our method against the physical commonsense score (1-5) from VideoPhy2 and VideoScore2, and compared against the Visual Quality (VQ) and Motion Quality (MQ) from VisionReward and VideoReward.

\textbf{Benchmarks.} As we are computing the pairwise accuracy, we reorder the existing benchmark datasets and form pairs by matching their correspondence, as shown in Table~\ref{tab:eval_data_all}. For perceptual quality benchmark, we adopt the existing benchmarks directly as they are already in pairwise form. Details 

\begin{table*}[!h]
\centering
\small
\setlength{\tabcolsep}{4pt}
\caption{
Evaluation benchmarks used in this work, covering both \textbf{physical plausibility} and \textbf{perceptual quality} under diverse settings of correspondence and pair types.
Physical benchmarks focus on consistency of dynamics, while quality benchmarks evaluate alignment with human preferences.
}
\label{tab:eval_data_all}
\begin{tabular}{l l r l l l}
\toprule
\textbf{Domain} & \textbf{Eval Dataset} & \textbf{Pairs} & \textbf{Type} & \textbf{Corres.} & \textbf{Desc.} \\
\midrule

\multirow{4}{*}{Physical}
& ImplausiBench~\citep{motamed2025travlrecipemakingvideolanguage} 
& 150 & R--G & Img+Txt 
& Paired with real and physical violations \\

& PhyDetEx~\citep{wang2025phydetexdetectingexplainingphysical} 
& 2K & R--G & None 
& Diverse real vs synthetic violations \\

& VideoPhy2-Test~\citep{bansal2025videophy2challengingactioncentricphysical} 
& 1.2K & G--G & Txt 
& Paired with score differences \\

& VideoFeedback2-Test~\citep{he2025videoscore2thinkscoregenerative} 
& 2K & G--G & None 
& Paired with score differences \\

\midrule

\multirow{4}{*}{Quality}
& MonetBench~\citep{xu2026visionrewardfinegrainedmultidimensionalhuman} 
& 1K & G--G & Txt 
& Human preference on video quality \\

& Rapidata-I2V~\citep{rapidata_hailuo02_marey_i2v_2025, rapidata_seedance1pro_i2v_2025} 
& 295 & G--G & Img+Txt 
& Human preference on video quality \\

& RewardBench~\citep{liu2025improvingvideogenerationhuman} 
& 25K & G--G & Txt 
& Human preference on video quality \\

\bottomrule
\end{tabular}
\end{table*}

\textbf{Human Correlation.} We reported the human correlation from the official VideoPhy2~\citep{bansal2025videophy2challengingactioncentricphysical} test set and VideoFeedback2~\citep{he2025videoscore2thinkscoregenerative} test set.

\subsection{Reproducibility}
\label{app:b_reproducibility}
To ensure reproducibility, all stochastic components of our pipeline are controlled via a fixed global seed of 42. All evaluation protocols, including correlation scoring and automated pairwise comparison are initialized with the same seed. For pairwise evaluation, we randomize the A/B presentation order to mitigate position bias; critically, all swap assignments are pre-computed from a seeded RNG prior to dispatching any worker threads, ensuring that results are invariant to worker count and non-deterministic task completion order.

\clearpage
\begin{figure*}[h]
\centering
\begin{tcolorbox}[title={Physical Consistency Pairwise Evaluation Prompt}]
\begin{verbatim}
You are evaluating the physical consistency of two generated videos. 
Physical consistency refers to whether objects move, deform, interact, 
and persist in ways that respect the constraints of the physical world.
Focus ONLY on physical consistency. Ignore visual quality, aesthetics, 
text rendering, or semantic coherence.

Physical violations include:
- objects defying gravity
- passing through surfaces
- impossible deformations
- discontinuities in motion
- violations of object permanence
- implausible contact or collision
- fluids/cloth/hair behaving impossibly

Video A:
<video>{video_a_base64}</video>
Video B:
<video>{video_b_base64}</video>

Which video is more physically consistent?
**Vote:** [A or B]
**Justification:** [1-2 sentences]
\end{verbatim}
\end{tcolorbox}
\caption{Prompt template for pairwise physical consistency evaluation between two generated videos.}
\label{fig:physical-consistency-pairwise-prompt}
\end{figure*}

\begin{figure*}[h]
\centering
\begin{tcolorbox}[title={Physical Consistency Scoring Prompt}]
\begin{verbatim}
You are evaluating the physical consistency of a generated video.
Physical consistency refers to whether objects move, deform,
interact, and persist in ways that respect the constraints of
the physical world. 
Watch this video carefully and evaluate its physical plausibility. 
Score the video on a scale from 1 to 5:
    1 - severe violations of physical laws
        (impossible motion, objects, forces)
    2 - major physical implausibilities
    3 - some implausible elements but mostly reasonable
    4 - minor implausibilities, mostly realistic physics
    5 - fully physically plausible, natural and realistic
        throughout

Respond with a JSON object containing a single key "score"
with an integer value 1-5.
\end{verbatim}
\end{tcolorbox}
\caption{Prompt template for scalar physical consistency scoring of a single video.}
\label{fig:physical-consistency-score-prompt}
\end{figure*}

\clearpage
\section{Data Study}

\textbf{Concerns of train / test overlap.} Supervision instances are disjoint between training and test sets for every benchmark (Table~\ref{tab:train_test_overlap} and~\ref{tab:official_overlap}). However, VideoFeedback2-test shares all prompts and generators with VideoFeedback2-train, so results on VF2 reflect held-out annotations under the same prompt and generator distribution rather than a distribution shift. VideoPhy2-test, by contrast, shares no prompts and only partially overlaps in generators.

\begin{table}[h]
\centering
\caption{
Train/test overlap between each evaluation benchmark and the complete training corpus used by \phyprobe{}.
Entries are reported as $X/Y$, where $X$ denotes the number of unique evaluation video-label entities that also appear in the complete training corpus and $Y$ denotes the total number of unique evaluation entities.
Supervision overlap is computed over annotated supervision instances (e.g., preference labels, quality scores, or pairwise comparisons).
Generation Prompt overlap is computed over unique conditioning inputs used to generate the evaluation videos (i.e., text prompts for text-to-video tasks and image--text input pairs for image-conditioned tasks).
Generator overlap is computed over unique video generation models. Generator overlap does not imply supervision or train--test sample overlap.
}
\label{tab:train_test_overlap}
\resizebox{\columnwidth}{!}{
\begin{tabular}{lccc}
\toprule
\textbf{Evaluation Benchmark} &
\textbf{Supervision overlap} &
\textbf{Generation Prompt overlap} &
\textbf{Generator overlap} \\
\midrule

\multicolumn{4}{l}{\textbf{Physical Domain}} \\
\midrule
 ImplB$^p$(ImplausiBench)       & 0/300   & Unavailable & Unavailable  \\
VP2$^p$ (VideoPhy2-test)      & 0/2888   & 0/599  & 2/5 (1172/2943 samples)  \\
PhyDEx$^p$ (PhyDetEx)            & 0/398   & 0/247 & 2/3 (55/250 samples)  \\
VF2$^p$ (VideoFeedback2-test) & 0/500 & 500/500 & 25/25 (500/500 samples) \\
\midrule

\multicolumn{4}{l}{\textbf{Quality Domain}} \\
\midrule
MonetBench          & 0/2000   & 2/2000 & Unavailable  \\
Rapidata-I2V        & 0/298   & 0/298 & 0/3 (0/298 samples) \\
RewardBench         & 0/4691   & Unavailable & 12/12 (4691/4691 samples)  \\
\bottomrule
\end{tabular}
}
\end{table}

\begin{table}[h]
\centering
\caption{
Official train/test overlap for the benchmark datasets used in \phyprobe{}.
Supervision overlap denotes identical annotated supervision instances shared between the official training and evaluation splits.
Generation Prompt overlap denotes identical conditioning inputs (i.e., text prompts or image--text input pairs).
Generator overlap denotes shared video generation models and is reported only when generator metadata is available. Note that VideoFeedback2's official splits share prompts and generators; supervision is disjoint, but the test set does not constitute a prompt- or generator-level distribution shift.
}
\label{tab:official_overlap}
\resizebox{\columnwidth}{!}{
\begin{tabular}{llccc}
\toprule
\textbf{Training} &
\textbf{Evaluation} &
\textbf{Supervision overlap} &
\textbf{Generation Prompt overlap} &
\textbf{Generator overlap} \\
\midrule
VideoPhy2-train      & VideoPhy2-test      & 0\%   & 0\% & 43\% \\
VideoFeedback2-train & VideoFeedback2-test & 0\% & 100\% & 100\% \\
TRAVL                & ImplausiBench       & 0\%   & 0\% & Unavailable \\
\bottomrule
\end{tabular}
}
\end{table}

\begin{figure}[!h]
    \centering

    \begin{subfigure}[t]{0.48\linewidth}
        \centering
        \includegraphics[width=\linewidth]{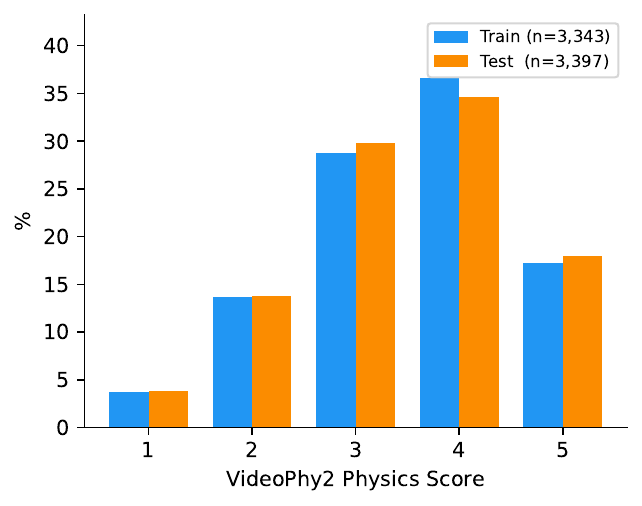}
        \caption{VideoPhy2 Annotations}
    \end{subfigure}
    \hfill
    \begin{subfigure}[t]{0.48\linewidth}
        \centering
        \includegraphics[width=\linewidth]{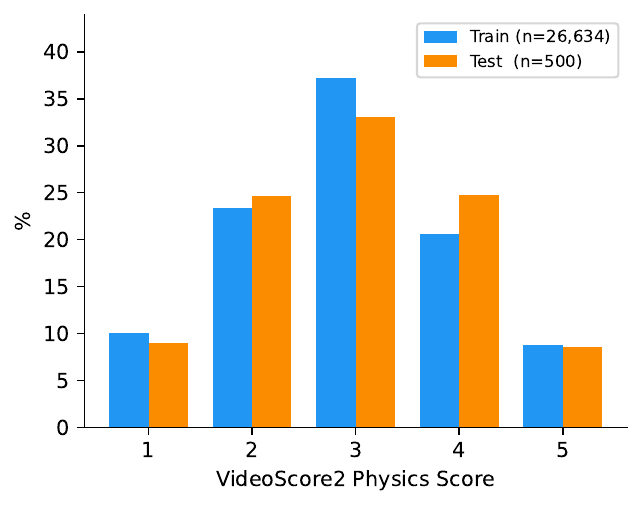}
        \caption{VideoScore2 Annotations}
    \end{subfigure}

    \caption{
    \textbf{Human annotation distributions across different physical consistency evaluation datasets.} Annotators consistently avoid extreme ratings and concentrate around intermediate scores, reflecting the coarse and thresholded nature of human perception toward physical inconsistencies. Anchor-based supervision serves as a mitigation strategy by providing stable reference points at the low- and high-violation ends of the spectrum, helping calibrate the absolute violation scale across heterogeneous datasets.
    }
    \vspace{-0.6cm}
    \label{fig:dataset_score_range}
\end{figure}

\textbf{Human Scoring Analysis in VideoPhy2 and VideoScore2.} In Figure~\ref{fig:dataset_score_range} We study the human annotations in both training and test data from VideoPhy2~\citep{bansal2025videophy2challengingactioncentricphysical} and VideoScore2~\citep{he2025videoscore2thinkscoregenerative}. We found that Annotators consistently avoid extreme ratings and concentrate around intermediate scores, reflecting the coarse and thresholded nature of human perception toward physical inconsistencies. Anchor-based supervision serves as a mitigation strategy by providing stable reference points at the low- and high-violation ends of the spectrum, helping calibrate the absolute violation scale across heterogeneous datasets.

\textbf{Heuristic Labels for BrokenVideos.}
The BrokenVideos dataset does not provide explicit pairwise preference annotations. We therefore construct heuristic preference labels based on automatically extracted anomaly statistics. Specifically, we compute a composite anomaly score from mask area, center proximity, and mask presence ratio (Table~\ref{tab:brokenvideos_statistics}). To improve label reliability for cross-prompt comparisons, we retain only pairs whose anomaly scores differ by at least 0.4. As shown in Table~\ref{tab:brokenvideos_score}, this threshold is approximately $2.9\times$ the interquartile range (IQR) of the score distribution, representing a conservative criterion that selects well-separated preference pairs. Finally, we validate the resulting heuristic labels through a human annotation study involving four independent annotators. Table~\ref{tab:heuristic_validation} shows that the heuristic agrees with the human majority on 80.6\% of non-tied pairs, while the observed heuristic--human agreement is comparable to the measured human--human agreement, supporting the use of the proposed heuristic as a reliable source of auxiliary supervision.

\begin{table}[h]
\centering
\small
\caption{Distribution of per-video statistics in the BrokenVideos training set. Lower mask area, center proximity, and mask presence ratio indicate fewer localized anomalies. Annotation coverage ratio is reported only as a data quality statistic and is not used in the composite anomaly score.}
\label{tab:brokenvideos_statistics}
\begin{tabular}{lrrrrrr}
\toprule
\textbf{Field} & \textbf{Min} & \textbf{Q1} & \textbf{Median} & \textbf{Q3} & \textbf{P99} & \textbf{Max} \\
\midrule
Mask area                 & 0.00011 & 0.014 & 0.039 & 0.094 & 0.446 & 0.860 \\
Center proximity          & 0.00    & 0.34  & 0.56  & 0.79  & 1.00  & 1.00  \\
Mask presence ratio       & 0.021   & 1.00  & 1.00  & 1.00  & 1.00  & 1.00  \\
\bottomrule
\end{tabular}
\end{table}

\begin{table}[t]
\centering
\small
\caption{Human validation of the proposed heuristic-derived cross-prompt BrokenVideos preference labels design, as the score is designed as
$0.6\cdot\text{mask\_area}
+0.3\cdot\text{center\_proximity}
+0.1\cdot\text{mask\_presence\_ratio}$. Four independent non-expert annotators evaluated the same set of 85 sampled video pairs.}
\label{tab:heuristic_validation}
\begin{tabular}{lc}
\toprule
\textbf{Metric} & \textbf{Value} \\
\midrule
\multicolumn{2}{l}{\textit{Heuristic vs. Human}} \\
Annotator 1 agreement & 77.6\% \\
Annotator 2 agreement & 74.1\% \\
Annotator 3 agreement & 77.6\% \\
Annotator 4 agreement & 65.9\% \\
Human majority agreement & 80.6\% (58/72) \\
\midrule
\multicolumn{2}{l}{\textit{Human vs. Human}} \\
Mean pairwise agreement & 77.5\% \\
Pairwise agreement range & 72--89\% \\
Fleiss' $\kappa$ & 0.42 \\
\midrule
\multicolumn{2}{l}{\textit{Unanimous Cases}} \\
4/4 unanimous pairs & 51 / 85 \\
Heuristic agreement on unanimous pairs & 86.3\% (44/51) \\
\bottomrule
\end{tabular}
\end{table}

\begin{table}[h]
\centering
\small
\caption{
Distribution of the composite anomaly score used for BrokenVideos pair construction.
The score is computed as
$0.6\cdot\text{mask\_area}
+0.3\cdot\text{center\_proximity}
+0.1\cdot\text{mask\_presence\_ratio}$.
The interquartile range (IQR) characterizes the intrinsic spread of the heuristic score.
The adopted cross-prompt threshold (0.4) is approximately $2.9\times$ the IQR, making it a conservative threshold that retains only well-separated preference pairs.
}
\label{tab:brokenvideos_score}
\begin{tabular}{cccccc}
\toprule
\textbf{Min} & \textbf{Q1} & \textbf{Median} & \textbf{Q3} & \textbf{IQR} & \textbf{Max} \\
\midrule
0.012 & 0.24 & 0.31 & 0.38 & 0.14 & 0.699 \\
\bottomrule
\end{tabular}
\end{table}

\clearpage

\section{Baseline Study}
\label{app:d_baseline_tie}

\textbf{Tie-aware analysis}. We further performed a tie-aware pairwise analysis of VideoPhy2-AutoEval and VideoScore2 (Table~\ref{tab:baseline_tieaware}). While VideoPhy2-AutoEval achieves higher accuracy after excluding tied pairs on several benchmarks, it predicts ties substantially more frequently than VideoScore2 across nearly all evaluation sets. This behavior is particularly pronounced on Rapidata-I2V, where VideoPhy2-AutoEval exhibits an 81.4\% tie rate, leaving relatively few non-tied comparisons and reducing its overall pairwise accuracy (9.5\% vs. 50.9\% on non-tied pairs). Rapidata-I2V also contains three recently released video generators that are absent from the training data of both VideoPhy2-AutoEval and VideoScore2, which may partially explain the particularly high tie rate observed on this benchmark.

\begin{table}[h]
\centering
\small
\setlength{\tabcolsep}{4pt}
\caption{\textbf{Tie-aware pairwise accuracy of VideoPhy2-AutoEval and VideoScore2 across evaluation benchmarks.} Both baselines output discrete physical-commonsense scores, so a pair receives a model tie whenever the two videos are assigned equal scores. \emph{Model tie rate} is the fraction of pairs on which this occurs. \emph{Acc (all pairs)} is computed over every pair, including human-labelled ties, with a model tie counted as correct only when the human label is also a tie; it corresponds to the ``w/ tie'' columns of Table~3. \emph{Acc (model non-tied)} restricts evaluation to pairs on which the model produced a non-tied prediction, i.e., accuracy conditional on the model committing to a preference.}
\begin{tabular}{llrrrr}
\toprule
\textbf{Eval Set} & \textbf{Model} & \textbf{Model tie rate} & \textbf{Acc (all pairs)} & \textbf{Acc (model non-tied)} \\
\midrule
\multirow{2}{*}{VP2$^p$}
  & VideoPhy2-AutoEval & 58.8\% & 28.5\% & 69.1\% \\
  & VideoScore2        & 40.2\% & 49.7\% & 60.0\% \\
\midrule
\multirow{2}{*}{VF2$^p$}
  & VideoPhy2-AutoEval & 46.0\% & 36.3\% & 67.2\% \\
  & VideoScore2        & 38.1\% & 65.0\% & 79.6\% \\
\midrule
\multirow{2}{*}{ImplB$^p$}
  & VideoPhy2-AutoEval & 63.3\% & 28.0\% & 76.4\% \\
  & VideoScore2        & 43.3\% & 68.7\% & 75.3\% \\
\midrule
\multirow{2}{*}{PhyDEx$^p$}
  & VideoPhy2-AutoEval & 40.0\% & 30.5\% & 50.8\% \\
  & VideoScore2        & 35.8\% & 60.6\% & 75.1\% \\
\midrule
\multirow{2}{*}{Rapidata-I2V}
  & VideoPhy2-AutoEval & \textbf{81.4\%} & 9.5\% & 50.9\% \\
  & VideoScore2        & 45.8\% & 45.8\% & 53.1\% \\
\midrule
\multirow{2}{*}{MonetBench}
  & VideoPhy2-AutoEval & 69.4\% & 22.5\% & 49.0\% \\
  & VideoScore2        & 51.2\% & 42.3\% & 46.7\% \\
\midrule
\multirow{2}{*}{RewardBench}
  & VideoPhy2-AutoEval & 64.1\% & 24.6\% & 54.9\% \\
  & VideoScore2        & 37.4\% & 57.4\% & 66.2\% \\
\bottomrule
\end{tabular}
\label{tab:baseline_tieaware}
\end{table}

\section{Societal Impact}
\label{app:societal}

Reliable evaluation of physical consistency in generated videos may help improve the safety, reliability, and realism of generative video systems by enabling better detection of implausible or unstable outputs. More robust physical evaluation could support applications in simulation, robotics, scientific visualization, education, and content generation, where physically inconsistent behavior may reduce trustworthiness or usability.

At the same time, advances in evaluation may indirectly contribute to improving the realism of generative models, including synthetic media that could be used for misinformation or deceptive content. However, \phyprobe{} is designed as an evaluation framework rather than a generation system, and does not itself generate or manipulate media. We also note that physical plausibility should not be conflated with factual correctness or authenticity: a physically plausible video may still depict fabricated events.

\section{Safeguards}
\label{app:safeguards}
\phyprobe{} is intended as a research evaluation framework for measuring physical consistency in generated videos, rather than a system for determining factual authenticity or detecting misinformation. We emphasize that physical plausibility should not be interpreted as evidence that a video depicts real events. In addition, \phyprobe{} remains imperfect under distribution shift and may assign incorrect scores to unusual but physically valid motions or complex long-horizon interactions. 

\clearpage



\end{document}